\pdfoutput=1

\documentclass[11pt]{article}

\usepackage[]{EMNLP2023}

\usepackage{times}
\usepackage{latexsym}
\usepackage{soul}
\usepackage{verbatim}
\usepackage{graphicx}
\usepackage{booktabs} 
\usepackage{multirow}
\usepackage{arydshln}
\usepackage{enumitem}
\usepackage{xcolor}
\usepackage{amsmath}
\usepackage{amssymb}
\usepackage{colortbl}
\usepackage{booktabs}
\usepackage{array}
\usepackage[most]{tcolorbox}
\usepackage{float}

\definecolor{customgreen}{rgb}{0.2, 0.7, 0.2}
\definecolor{basegreen}{rgb}{0.92, 1, 0.92}
\definecolor{customred}{rgb}{1, 0.2, 0.2}
\definecolor{basered}{rgb}{1, 0.95, 0.95}

\usepackage[T1]{fontenc}

\usepackage{soul}
\usepackage[utf8]{inputenc}

\usepackage{microtype}

\usepackage{inconsolata}

\newtcbox{\hlgradientgreen}[1][100]{on line, rounded corners, box align=base, colback=customgreen!#1!basegreen, colframe=white, size=fbox, arc=3pt, before upper=\strut, top=-2pt, bottom=-4pt, left=-2pt, right=-2pt, boxrule=0pt}
\newtcbox{\hlgradientred}[1][100]{on line, rounded corners, box align=base, colback=customred!#1!basered, colframe=white, size=fbox, arc=3pt, before upper=\strut, top=-2pt, bottom=-4pt, left=-2pt, right=-2pt, boxrule=0pt}

\newcommand{\colorcell}[2]{%
    \hspace{-8pt}%
    \pgfmathsetmacro{\base}{#2}
    \pgfmathsetmacro{\value}{#1}
    \pgfmathsetmacro{\diff}{\value - \base}
    \ifdim \diff pt > 0pt
        \pgfmathsetmacro{\percent}{min(\diff/(\base+0.01)*1500, 100)}
        \hlgradientgreen[\percent]{\raisebox{0pt}[0em][0em]{#1}}
    \else
        \ifdim \diff pt < 0pt
            \pgfmathsetmacro{\percent}{min(-\diff/(\base+0.01)*300, 100)}
            \hlgradientred[\percent]{\raisebox{0pt}[0em][0em]{#1}}
        \else
            #1
        \fi       
    \fi
}

\definecolor{battleshipgrey}{rgb}{0.3, 0.3, 0.3}
\definecolor{brilliantrose}{rgb}{1.0, 0.33, 0.64}
\definecolor{americanrose}{rgb}{1.0, 0.01, 0.24}
\definecolor{jweigreen}{rgb}{0,0.45,0.24}
\definecolor{bluegray}{rgb}{0.1, 0.1, 0.4}
\definecolor{ao(english)}{rgb}{0.0, 0.5, 0.0}
\definecolor{blanchedalmond}{rgb}{1.0, 0.92, 0.8}
\definecolor{atomictangerine}{rgb}{1.0, 0.6, 0.4}
\definecolor{chocolate(web)}{rgb}{0.82, 0.41, 0.12}
\definecolor{bananayellow}{rgb}{1.0, 0.88, 0.21}
\definecolor{goldenbrown}{rgb}{0.6, 0.4, 0.08}
\definecolor{aliceblue}{rgb}{0.94, 0.97, 1.0}
\definecolor{beige}{rgb}{0.96, 0.96, 0.86}
\definecolor{babyblue}{rgb}{0.54, 0.81, 0.94}
\definecolor{camel}{rgb}{0.76, 0.6, 0.42}
\definecolor{cinnamon}{rgb}{0.82, 0.41, 0.12}
\definecolor{deepskyblue}{rgb}{0.0, 0.75, 1.0}
\definecolor{frenchblue}{rgb}{0.0, 0.45, 0.73}
\definecolor{classicrose}{rgb}{0.98, 0.8, 0.91}
\definecolor{frenchrose}{rgb}{0.96, 0.29, 0.54}
\definecolor{frenchlilac}{rgb}{0.53, 0.38, 0.56}
\definecolor{frenchbeige}{rgb}{0.65, 0.48, 0.36}

\usepackage{arydshln}
\makeatletter
\def\adl@drawiv#1#2#3{%
        \hskip.5\tabcolsep
        \xleaders#3{#2.5\@tempdimb #1{1}#2.5\@tempdimb}%
                #2\z@ plus1fil minus1fil\relax
        \hskip.5\tabcolsep}
\newcommand{\cdashlinelr}[1]{%
  \noalign{\vskip 1.3pt
           \global\let\@dashdrawstore\adl@draw
           \global\let\adl@draw\adl@drawiv}
  \cdashline{#1}[.4pt/2pt]
  \noalign{\global\let\adl@draw\@dashdrawstore
           \vskip 1.3pt}}
\makeatother

\title{Is Preference Alignment Always the Best Option to Enhance LLM-Based Translation? An Empirical Analysis}

\author{Hippolyte Gisserot-Boukhlef$^{1,4}$ \quad Ricardo Rei$^{2}$ \quad Emmanuel Malherbe$^{1}$ \\  \textbf{Céline Hudelot}$^{4}$ \quad \textbf{Pierre Colombo}$^{3,4}$ \quad \textbf{Nuno M. Guerreiro}$^{2,4,5,6}$ \\
$^{1}$Artefact Research Center \quad $^2$Unbabel \quad $^3$Equall \\ 
$^4$MICS, CentraleSupélec, Université Paris-Saclay \quad $^5$Instituto de Telecomunicações
\\
$^6$Instituto Superior Técnico \& Universidade de Lisboa (Lisbon ELLIS Unit)
\\
\small \url{hippolyte.gisserot-boukhlef@centralesupelec.fr}}

\begin{document}
\maketitle

\begin{abstract}

Neural metrics for machine translation (MT) evaluation have become increasingly prominent due to their superior correlation with human judgments compared to traditional lexical metrics. Researchers have therefore utilized neural metrics through quality-informed decoding strategies, achieving better results than likelihood-based methods. With the rise of Large Language Models (LLMs), preference-based alignment techniques have gained attention for their potential to enhance translation quality by optimizing model weights directly on preferences induced by quality estimators. This study focuses on Contrastive Preference Optimization (CPO) and conducts extensive experiments to evaluate the impact of preference-based alignment on translation quality. Our findings indicate that while CPO consistently outperforms Supervised Fine-Tuning (SFT) on high-quality data with regard to the alignment metric, it may lead to instability across downstream evaluation metrics, particularly between neural and lexical ones. Additionally, we demonstrate that relying solely on the base model for generating candidate translations achieves performance comparable to using multiple external systems, while ensuring better consistency across downstream metrics.\footnote{All relevant preference datasets and aligned models, along with detailed evaluation metrics, are available at \href{https://huggingface.co/collections/artefactory/translation-alignment-analysis-66f3e56669bed67108c309ea}{\texttt{https://huggingface.co/collections/artefactory/ translation-alignment-analysis}}.}

\end{abstract}

\section{Introduction}

Neural metrics for machine translation evaluation that are trained to mimic human preferences, such as BLEURT \cite{sellam-etal-2020-bleurt}, COMET \cite{rei-etal-2020-comet, rei-etal-2022-comet}, or Metric-X \cite{juraska-etal-2023-metricx}, have become increasingly prevalent. These metrics offer greater accuracy and better reflect human judgments compared to traditional lexical metrics \cite{mathur-etal-2020-tangled, kocmi-etal-2021-ship, freitag-etal-2022-results, kocmi2024navigating} like BLEU \cite{papineni2002bleu}, METEOR \cite{banerjee-lavie-2005-meteor} or chrF \cite{popovic2015chrf}, which mainly consider lexical overlap with a reference text. As such, researchers have attempted to leverage these improvements by integrating them directly into translation systems. 

One appealing strategy to incorporate quality information to improve downstream translation performance involves using decoding strategies such as N-Best reranking and Minimum Bayes Risk (MBR) decoding \cite{kumar2002minimum, kumar-byrne-2004-minimum, eikema-aziz-2020-map,  fernandes-etal-2022-quality, freitag-etal-2022-high}. These techniques rely on generating multiple candidates to maximize a given quality metric at inference time, and research has shown that they consistently yield better results than likelihood-based decoding techniques \cite{eikema-aziz-2020-map, koehn-knowles-2017-six, pmlr-v80-ott18a}.

With the rise of decoder-only LLMs in MT, quality-informed fine-tuning techniques have gained significant attention. Unlike decoding-based methods that inject quality information at inference time, fine-tuning modifies model weights using training sets induced with quality information. These approaches include filtering parallel training data based on a quality metric \cite{alves2024tower}, distilling gains from more expensive quality-aware methods such as MBR~\cite{finkelstein2024mbrqefinetuningtrainingtime}, or employing preference-based alignment techniques \cite{rafailov2024direct, xu2024contrastive}, where the model learns preferences induced by quality metrics between candidate translations typically generated by multiple systems. In this work, we focus specifically on the latter.

Alignment techniques represent a paradigm shift from quality-aware inference time approaches, as they optimize the metric of interest \textit{indirectly}. Understanding the impact of these approaches on translation quality is thus a relevant problem. While some studies have examined quality-informed decoding techniques and their influence on translation output~\cite{amrhein2022identifying}, there is still a gap in our understanding of how preference-based fine-tuning affects translation quality.

In this work, we aim to bridge this gap by examining the properties of preference-based alignment techniques, with a particular focus on Contrastive Preference Optimization (CPO) \cite{xu2024contrastive}, which has been used successfully to achieve very competitive translation performance. Our analysis seeks to describe the effects of preference-based fine-tuning on downstream performance, specifically regarding alignment effectiveness, the interactions between optimized and non-optimized metrics, and the impact of using multiple candidate translation systems for generating preference data. Through extensive experimentation, we find that:

\begin{itemize}
    \item Preference-based alignment globally outperforms Supervised Fine-Tuning (SFT) on high-quality data in terms of maximizing the alignment metric.

    \item However, preference-based alignment is highly sensitive to the choice of candidate systems used for generating preference data, affecting both the alignment metric and downstream metric consistency.
    
    \item Aligning a model using its own translations achieves performance comparable to employing multiple external systems, while ensuring better metric consistency and allowing for improved control over the alignment process.
\end{itemize}

\section{Background}

\subsection{Quality-Informed Translation}

Along with human evaluation, lexical metrics like BLEU \cite{papineni2002bleu}, chrF \cite{popovic2015chrf}, METEOR \cite{banerjee-lavie-2005-meteor}, and ROUGE \cite{lin-2004-rouge} have long been used for translation evaluation. However, human evaluation is costly, and lexical metrics have been shown to correlate poorly with human judgements.

More recently, some neural metrics have emerged as a preferred method to mimic human preferences without relying on expensive human evaluation. The intuitive approach involves training an encoder model on human-annotated source-translation-reference triplets. Among the metrics most frequently mentioned in the literature are BLEURT \cite{yan-etal-2023-bleurt}, COMET \cite{rei-etal-2020-comet}, CometKiwi \cite{rei-etal-2022-cometkiwi}, xCOMET \cite{guerreiro2023xcomet}, and Metric-X \cite{juraska-etal-2023-metricx}. They can be divided into two families: \textit{reference-based} metrics, that include a human-written gold reference as an input to the scoring model, and \textit{reference-free} metrics, which only require access to the source sentence and the generated translation. These neural metrics have proven particularly effective at scoring translations and achieve much higher correlation with human judgments than their lexical counterparts \cite{mathur-etal-2020-tangled, kocmi-etal-2021-ship, freitag-etal-2022-results, kocmi2024navigating}.

These neural metrics have also been leveraged to improve translation models through decoding strategies. The approach involves sampling various candidate translations, scoring them according to a given metric, and selecting the one with the highest score. This methodology is exemplified by MBR decoding in the reference-based setting and N-best reranking in the reference-free setting \cite{fernandes-etal-2022-quality, freitag-etal-2022-high}.

\subsection{Quality-Based Fine-Tuning}

With the recent rise of decoder-only LLMs applied to translation tasks \cite{zhu2023multilingual, jiao2023chatgpt, hendy2023good, kocmi-etal-2023-findings, freitag-etal-2023-results, xu2023paradigm,alves2023steering, xu2024contrastive, alves2024tower}, and with automatic metrics increasingly reflecting human judgments \cite{sellam-etal-2020-bleurt, rei-etal-2020-comet, juraska-etal-2023-metricx}, quality-based fine-tuning has gained considerable traction. This approach shifts the objective from selecting the best candidate translation according to a metric at inference time to directly updating model weights through fine-tuning to produce the desired translations. A straightforward approach is to perform SFT on high-quality translations, evaluated and then filtered with respect to a metric of interest \cite{alves2024tower}.

Another attractive alternative is Preference Optimization (PO) \cite{simianer2018preference, rafailov2024direct, xu2024contrastive, yang2023direct, xu2024advancing, wu2024word}, which focuses on learning preferences between chosen and rejected translations rather than simply increasing the likelihood of high-quality sentences. A popular PO method is Direct Preference Optimization (DPO) \cite{rafailov2024direct}, which aims to maximize a scaled likelihood gap between a chosen and a rejected option. More recently, CPO \cite{xu2024contrastive} has emerged as a promising alternative, incorporating an SFT term into the DPO loss, effectively combining the strengths of both methods. Moreover, by removing the reference policy from the learning objective, it improves training efficiency.

\section{Experimental Setup}
\label{sec:xp_setup}

\begin{table*}[t]
\renewcommand\arraystretch{0.9}
\centering
\small
\resizebox{0.99\textwidth}{!}
{\begin{tabular}{lccccccccc}
\toprule
 & \multicolumn{4}{c}{\texttt{xx-en}} &  & \multicolumn{4}{c}{\texttt{en-xx}} \\
\noalign{\vskip 0.5pt}
\cline{2-5} \cline{7-10}
\noalign{\vskip 2pt}
 & \multicolumn{2}{c}{Neural} &  & Lexical &  & \multicolumn{2}{c}{Neural} &  & Lexical \\
\noalign{\vskip 0.5pt}
\cline{2-3} \cline{5-5} \cline{7-8} \cline{10-10}\noalign{\vskip 2pt}
 & xCOMET-QE & CometKiwi &  & chrF &  & xCOMET-QE & CometKiwi &  & chrF \\
\midrule
\textbf{Base} & \textcolor{white}{$\bullet$} \colorcell{87.80}{87.80} & \textcolor{white}{$\bullet$} \colorcell{80.86}{80.86} &  & \textcolor{white}{$\bullet$} \colorcell{58.53}{58.53} &  & \textcolor{white}{$\bullet$} \colorcell{91.91}{91.91} & \textcolor{white}{$\bullet$} \colorcell{81.17}{81.17} &  & \textcolor{white}{$\bullet$} \colorcell{49.49}{49.49} \\
\midrule
\multicolumn{10}{c}{\textit{Preferences induced with} \textbf{xCOMET-QE}} \\\cdashlinelr{1-10}
SFT & \textcolor{atomictangerine!60}{$\bullet$} \colorcell{89.13}{87.80} & \textcolor{white}{$\bullet$} \colorcell{81.49}{80.86} &  & \textcolor{white}{$\bullet$} \colorcell{59.82}{58.53} &  & \textcolor{atomictangerine!60}{$\bullet$} \colorcell{92.38}{91.91} & \textcolor{white}{$\bullet$} \colorcell{81.67}{81.17} &  & \textcolor{white}{$\bullet$} \colorcell{50.28}{49.49} \\
CPO & \textcolor{atomictangerine!60}{$\bullet$} \textit{\colorcell{89.95}{87.80}} & \textcolor{white}{$\bullet$} \textit{\colorcell{81.89}{80.86}} &  & \textcolor{white}{$\bullet$} \colorcell{59.83}{58.53} &  & \textcolor{atomictangerine!60}{$\bullet$} \textit{\colorcell{92.75}{91.91}} & \textcolor{white}{$\bullet$} \textit{\colorcell{83.60}{81.17}} &  & \textcolor{white}{$\bullet$} \textit{\colorcell{47.69}{49.49}} \\
\midrule
\multicolumn{10}{c}{\textit{Preferences induced with} \textbf{CometKiwi}} \\\cdashlinelr{1-10}
SFT & \textcolor{white}{$\bullet$} \colorcell{89.26}{87.80} & \textcolor{atomictangerine!60}{$\bullet$} \colorcell{81.70}{80.86} &  & \textcolor{white}{$\bullet$} \colorcell{60.01}{58.53} &  & \textcolor{white}{$\bullet$} \colorcell{92.44}{91.91} & \textcolor{atomictangerine!60}{$\bullet$} \colorcell{81.93}{81.17} &  & \textcolor{white}{$\bullet$} \colorcell{50.49}{49.49} \\
CPO & \textcolor{white}{$\bullet$} \textit{\colorcell{89.82}{87.80}} & \textcolor{atomictangerine!60}{$\bullet$} \textit{\colorcell{82.04}{80.86}} &  & \textcolor{white}{$\bullet$} \textit{\colorcell{60.22}{58.53}} &  & \textcolor{white}{$\bullet$} \textit{\colorcell{92.19}{91.91}} & \textcolor{atomictangerine!60}{$\bullet$} \textit{\colorcell{83.64}{81.17}} &  & \textcolor{white}{$\bullet$} \textit{\colorcell{48.11}{49.49}} \\
\midrule
\multicolumn{10}{c}{\textit{Preferences induced with} \textbf{chrF}} \\\cdashlinelr{1-10}
SFT & \textcolor{white}{$\bullet$} \colorcell{87.61}{87.80} & \textcolor{white}{$\bullet$} \colorcell{80.82}{80.86} &  & \textcolor{atomictangerine!60}{$\bullet$} \colorcell{56.97}{58.53} &  & \textcolor{white}{$\bullet$} \colorcell{92.20}{91.91} & \textcolor{white}{$\bullet$} \colorcell{81.70}{81.17} &  & \textcolor{atomictangerine!60}{$\bullet$} \colorcell{50.30}{49.49} \\
CPO & \textcolor{white}{$\bullet$} \textit{\colorcell{78.51}{87.80}} & \textcolor{white}{$\bullet$} \textit{\colorcell{75.62}{80.86}} &  & \textcolor{atomictangerine!60}{$\bullet$} \textit{\colorcell{45.32}{58.53}} &  & \textcolor{white}{$\bullet$} \textit{\colorcell{88.89}{91.91}} & \textcolor{white}{$\bullet$} \textit{\colorcell{80.99}{81.17}} &  & \textcolor{atomictangerine!60}{$\bullet$} \textit{\colorcell{42.50}{49.49}} \\
\bottomrule
\end{tabular}}
\caption{Comparison between SFT on preferred translations and CPO in the multi-system setting, using xCOMET-QE, CometKiwi and chrF as alignment metrics. The same $3$ metrics are reported for evaluation, separately for into-English (\texttt{xx-en}) and out-of-English (\texttt{en-xx}) translations on the WMT'22 dataset. \hlgradientgreen[50]{Green} shades indicate metric improvements over the base model, while \hlgradientred[50]{red} shades indicate metric decreases. We represent with (\textcolor{atomictangerine!60}{$\bullet$}) scenarios where the preference metric matches the evaluation metric. Values in \textit{italic} font denote statistically significant differences between SFT- and CPO-based alignment at the $5\%$ level, based on one-tailed paired Student's $t$-tests.}
\label{tab:results}
\end{table*}

Here, we detail our experimental setup, explaining how we built the preference data, and train and evaluate the models.

\subsection{Preference Data}
\label{sec:pref_data}

\paragraph{Preference datasets.}
To build a preference dataset, one needs candidate translations, an evaluation metric $m$ to score these translations, and a method to select chosen and rejected hypotheses. We denote a candidate dataset by
\begin{equation*}
\begin{split}
\mathcal{D} = \left\{ \left( x_i, \mathcal{Y}_i \right) \right\}_{i=1}^N
\end{split},
\end{equation*}
where $x_i$ denotes the source sentence and $\mathcal{Y}_i$ is a set of candidate translations. One can then derive a preference dataset, 
\begin{equation*}
\begin{split}
\mathcal{D}_{pref} = \left\{ \left( x_i, y_i^{r}, y_i^{c} \right) \right\}_{i=1}^N
\end{split},
\end{equation*}
where $y_i^{c} \in \mathcal{Y}_i$ (chosen hypothesis) is a translation preferred to $y_i^{r} \in \mathcal{Y}_i$ (rejected hypothesis) according to a metric $m$ and a given selection method. 

\paragraph{Multi-system approach.}
In the multi-system scenario, we follow the setting outlined by \citet{xu2024contrastive}. Candidate translations are generated using three different systems, namely ALMA-13B-LoRA (the base model we aim to align, referred to as Base) \cite{xu2023paradigm}, GPT-4 \cite{openai2023gpt}, and the human-written gold reference (referred to as Ref). Formally, for all data samples,
\begin{equation*}
\begin{split}
\mathcal{Y}_i^{multi} = \left\{ y_i^{Ref}, y_i^{Base}, y_i^{GPT\text{-}4} \right\}
\end{split}.
\end{equation*}
Then, for each sample, the three translations are evaluated with regard to $m$. The one with highest (resp. lowest) score is selected as the chosen (resp. rejected) hypothesis. Formally,
\begin{equation*}
    y_i^c = \underset{y \in \mathcal{Y}_i^{multi}} {\arg \max} \ m \left( y \right) \, \land\, 
            y_i^r = \underset{y \in \mathcal{Y}_i^{multi}}{\arg \min} \ m \left( y \right)
\end{equation*}

\paragraph{Mono-system approach.}

In the mono-system setting, we solely rely on the base model for candidate generation. For each source sentence, $K = 50$ candidates are top-$p$-sampled $(p = 0.6)$ with a temperature $\tau = 0.9$,\footnote{These are the default parameters used in the ALMA paper~\citep{xu2023paradigm, xu2024contrastive}.} and are then ranked based on evaluation metric $m$. For all samples, this results in a set of candidates
\begin{equation*}
\begin{split}
\mathcal{Y}_i^{mono} = \left\{ y_i^{1}, \cdots, y_i^{K} \right\}
\end{split},
\end{equation*}
where $y_i^{1} \preceq \cdots \preceq y_i^{K}$ are sorted in increasing quality order, with no loss of generality. Preference pairs are then derived to ensure that $y_i^r \preceq y_i^{Base} \preceq y_i^c$ holds for all samples. Further details on the construction of mono-system preference datasets are given in Section~\ref{sec:mono_system} and Appendix \ref{apdx:pref_build}.

\paragraph{Source dataset.}

We rely on the FLORES-200-based \cite{nllb2022no} dataset used in \citet{xu2024contrastive} as a primary data source. It includes over $20000$ translation pairs spanning six languages (English (\texttt{en}), Czech (\texttt{cs}), German (\texttt{de}), Icelandic (\texttt{is}), Russian (\texttt{ru}), and Chinese (\texttt{zh})) and covering ten language directions, either into-English (\texttt{xx-en}) or out-of-English (\texttt{en-xx}).

\paragraph{Alignment metrics.}
In line with \citet{xu2024contrastive}, we rely on reference-free neural metrics, namely xCOMET-QE-XXL \cite{guerreiro2023xcomet} (referred to as xCOMET-QE), and the WMT'23 version of CometKiwi-XXL \cite{rei-etal-2023-scaling} (denoted by CometKiwi), as well as on a reference-based lexical metric, chrF \cite{popovic2015chrf}. 

\subsection{Training}

\textbf{Learning objective.}
We focus our diagnosis on CPO \cite{xu2024contrastive}, which combines a preference term with a likelihood term and achieves state-of-the-art performance in preference-based metric alignment for translation tasks. The empirical loss function is formally expressed as:
\begin{equation*}
    \begin{split}
    \mathcal{L}_{CPO} =& - \frac{1}{N} \sum_{i=1}^N \left[ \log \sigma \left( \beta \log \frac{\pi_{\theta} \left( y_i^c | x_i \right)}{\pi_{\theta} \left( y_i^r | x_i \right)} \right) \right] \\
    &+ \mathcal{L}_{SFT},
    \end{split}
\end{equation*}
where $\mathcal{L}_{SFT} = - \frac{1}{N} \sum_{i=1}^N \left[ \log \pi_{\theta} \left( y_i^c | x_i \right) \right]$ is the negative-log-likelihood loss applied to chosen translations, $\pi_{\theta}$ is the model to fine-tune, $\sigma$ is the sigmoid function and $\beta$ is a hyperparameter. In our experiments, CPO alignment is consistently compared to vanilla SFT on chosen translations.\footnote{All our models are trained using the code implementation provided by \citet{xu2024contrastive}.}

\paragraph{Training parameters.}
We replicate the exact same parameters as the ones outlined by \citet{xu2024contrastive}. ALMA-13B-LoRA is LoRA fine-tuned with rank $16$ for one epoch, starting with a learning rate of $10^{-4}$, using inverse square root decay and a batch size of $128$. The $\beta$ parameter of the CPO objective function is set equal to $0.1$, in line with the original DPO paper by \citet{rafailov2024direct}.

\subsection{Evaluation}

\paragraph{Inference setup.}

Following other works on LLM-based translation \cite{alves2024tower, briakou2024translating}, all generations at inference time are produced using greedy decoding, as it provides maximum computational efficiency while preserving high output quality.\footnote{Inference is performed using the vLLM library \cite{kwon2023efficient}.}

\paragraph{Evaluation datasets.} 
We evaluate our approaches on the WMT'22 test dataset, which consists of $17471$ source-reference pairs and includes the same ten language pairs as the preference data. Evaluations on WMT'23 test data are provided in Appendix \ref{apdx:results}.

\paragraph{Evaluation metrics.}
We use the same three metrics used to create the preference datasets: xCOMET-QE, CometKiwi, and chrF. Additional evaluation metrics are reported in Appendix \ref{apdx:results}, specifically the reference-based version of Metric-X-Large (referred to as Metric-X) \cite{juraska-etal-2023-metricx}, and BLEU \cite{papineni2002bleu}.

\section{Multi-System Preference Fine-Tuning}

We begin our analysis by focusing on the multi-system setting \cite{xu2024contrastive}, in which the chosen and rejected options are derived from a pool of three candidate systems consisting of ALMA-13B-LoRA (base model), GPT-4, and the gold reference.

\subsection{Top-Level Analysis}
\label{sec:top_level}

\begin{table*}[t]
\renewcommand\arraystretch{0.9}
\centering
\small
\resizebox{0.99\textwidth}{!}
{\begin{tabular}{lccccccccc}
\toprule
  & \multicolumn{4}{c}{\texttt{xx-en}} &  & \multicolumn{4}{c}{\texttt{en-xx}} \\
\noalign{\vskip 0.5pt}
\cline{2-5}
\cline{7-10}
\noalign{\vskip 2pt}
 & \multicolumn{2}{c}{Neural} &  & Lexical &  & \multicolumn{2}{c}{Neural} &  & Lexical \\
\noalign{\vskip 0.5pt}
\cline{2-3}
\cline{5-5}
\cline{7-8}
\cline{10-10}
\noalign{\vskip 2pt}
 & xCOMET-QE & CometKiwi &  & chrF &  & xCOMET-QE & CometKiwi &  & chrF \\
\midrule
\textbf{Base} & \textcolor{white}{$\bullet$} \colorcell{87.80}{87.80} & \textcolor{white}{$\bullet$} \colorcell{80.86}{80.86} &  & \textcolor{white}{$\bullet$} \colorcell{58.53}{58.53} &  & \textcolor{white}{$\bullet$} \colorcell{91.91}{91.91} & \textcolor{white}{$\bullet$} \colorcell{81.17}{81.17} &  & \textcolor{white}{$\bullet$} \colorcell{49.49}{49.49} \\\midrule
\multicolumn{10}{c}{\textit{Optimization via} \textbf{SFT}} \\\cdashlinelr{1-10}
\multicolumn{10}{l}{\textit{Preferences induced with} \textbf{xCOMET-QE}}\\
All systems & \textcolor{atomictangerine!60}{$\bullet$} \colorcell{89.13}{87.80} & \textcolor{white}{$\bullet$} \colorcell{81.49}{80.86} &  & \textcolor{white}{$\bullet$} \colorcell{59.82}{58.53} &  & \textcolor{atomictangerine!60}{$\bullet$} \colorcell{92.38}{91.91} & \textcolor{white}{$\bullet$} \colorcell{81.67}{81.17} &  & \textcolor{white}{$\bullet$} \colorcell{50.28}{49.49} \\
No Base & \textcolor{atomictangerine!60}{$\bullet$} \textit{\colorcell{89.41}{87.80}} & \textcolor{white}{$\bullet$} \colorcell{81.56}{80.86} &  & \textcolor{white}{$\bullet$} \textit{\colorcell{60.26}{58.53}} &  & \textcolor{atomictangerine!60}{$\bullet$} \colorcell{92.32}{91.91} & \textcolor{white}{$\bullet$} \colorcell{81.65}{81.17} &  & \textcolor{white}{$\bullet$} \textit{\colorcell{50.52}{49.49}} \\
No Ref & \textcolor{atomictangerine!60}{$\bullet$} \textit{\colorcell{89.32}{87.80}} & \textcolor{white}{$\bullet$} \textit{\colorcell{81.58}{80.86}} &  & \textcolor{white}{$\bullet$} \textit{\colorcell{60.08}{58.53}} &  & \textcolor{atomictangerine!60}{$\bullet$} \textit{\colorcell{92.22}{91.91}} & \textcolor{white}{$\bullet$} \textit{\colorcell{81.33}{81.17}} &  & \textcolor{white}{$\bullet$} \textit{\colorcell{50.05}{49.49}} \\
No GPT-4 & \textcolor{atomictangerine!60}{$\bullet$} \textit{\colorcell{88.44}{87.80}} & \textcolor{white}{$\bullet$} \textit{\colorcell{81.15}{80.86}} &  & \textcolor{white}{$\bullet$} \textit{\colorcell{58.86}{58.53}} &  & \textcolor{atomictangerine!60}{$\bullet$} \colorcell{92.33}{91.91} & \textcolor{white}{$\bullet$} \colorcell{81.74}{81.17} &  & \textcolor{white}{$\bullet$} \textit{\colorcell{50.06}{49.49}} \\\midrule
\multicolumn{10}{l}{\textit{Preferences induced with} \textbf{chrF}}\\
All systems & \textcolor{white}{$\bullet$} \colorcell{87.61}{87.80} & \textcolor{white}{$\bullet$} \colorcell{80.82}{80.86} &  & \textcolor{atomictangerine!60}{$\bullet$} \colorcell{56.97}{58.53} &  & \textcolor{white}{$\bullet$} \colorcell{92.20}{91.91} & \textcolor{white}{$\bullet$} \colorcell{81.70}{81.17} &  & \textcolor{atomictangerine!60}{$\bullet$} \colorcell{50.30}{49.49} \\
No Ref & \textcolor{white}{$\bullet$} \textit{\colorcell{89.21}{87.80}} & \textcolor{white}{$\bullet$} \textit{\colorcell{81.49}{80.86}} &  & \textcolor{atomictangerine!60}{$\bullet$} \textit{\colorcell{60.17}{58.53}} &  & \textcolor{white}{$\bullet$} \colorcell{91.99}{91.91} & \textcolor{white}{$\bullet$} \textit{\colorcell{80.96}{81.17}} &  & \textcolor{atomictangerine!60}{$\bullet$} \textit{\colorcell{50.57}{49.49}} \\\midrule\midrule
\multicolumn{10}{c}{\textit{Optimization via} \textbf{CPO}} \\\cdashlinelr{1-10}
\multicolumn{10}{l}{\textit{Preferences induced with} \textbf{xCOMET-QE}} \\
All systems & \textcolor{atomictangerine!60}{$\bullet$} \colorcell{89.95}{87.80} & \textcolor{white}{$\bullet$} \colorcell{81.89}{80.86} &  & \textcolor{white}{$\bullet$} \colorcell{59.83}{58.53} &  & \textcolor{atomictangerine!60}{$\bullet$} \colorcell{92.75}{91.91} & \textcolor{white}{$\bullet$} \colorcell{83.60}{81.17} &  & \textcolor{white}{$\bullet$} \colorcell{47.69}{49.49} \\
No Base & \textcolor{atomictangerine!60}{$\bullet$} \textit{\colorcell{89.59}{87.80}} & \textcolor{white}{$\bullet$} \textit{\colorcell{81.73}{80.86}} &  & \textcolor{white}{$\bullet$} \textit{\colorcell{59.94}{58.53}} &  & \textcolor{atomictangerine!60}{$\bullet$} \colorcell{92.74}{91.91} & \textcolor{white}{$\bullet$} \textit{\colorcell{83.13}{81.17}} &  & \textcolor{white}{$\bullet$} \textit{\colorcell{48.54}{49.49}} \\
No Ref & \textcolor{atomictangerine!60}{$\bullet$} \colorcell{89.91}{87.80} & \textcolor{white}{$\bullet$} \colorcell{81.86}{80.86} &  & \textcolor{white}{$\bullet$} \textit{\colorcell{60.59}{58.53}} &  & \textcolor{atomictangerine!60}{$\bullet$} \textit{\colorcell{92.44}{91.91}} & \textcolor{white}{$\bullet$} \textit{\colorcell{81.97}{81.17}} &  & \textcolor{white}{$\bullet$} \textit{\colorcell{50.67}{49.49}} \\
No GPT-4 & \textcolor{atomictangerine!60}{$\bullet$} \textit{\colorcell{88.81}{87.80}} & \textcolor{white}{$\bullet$} \textit{\colorcell{81.35}{80.86}} &  & \textcolor{white}{$\bullet$} \textit{\colorcell{57.91}{58.53}} &  & \textcolor{atomictangerine!60}{$\bullet$} \textit{\colorcell{92.22}{91.91}} & \textcolor{white}{$\bullet$} \textit{\colorcell{83.16}{81.17}} &  & \textcolor{white}{$\bullet$} \textit{\colorcell{46.82}{49.49}} \\\midrule
\multicolumn{10}{l}{\textit{Preferences induced with} \textbf{chrF}} \\
All systems & \textcolor{white}{$\bullet$} \colorcell{78.51}{87.80} & \textcolor{white}{$\bullet$} \colorcell{75.62}{80.86} &  & \textcolor{atomictangerine!60}{$\bullet$} \colorcell{45.32}{58.53} &  & \textcolor{white}{$\bullet$} \colorcell{88.89}{91.91} & \textcolor{white}{$\bullet$} \colorcell{80.99}{81.17} &  & \textcolor{atomictangerine!60}{$\bullet$} \colorcell{42.50}{49.49} \\
No Ref & \textcolor{white}{$\bullet$} \textit{\colorcell{89.26}{87.80}} & \textcolor{white}{$\bullet$} \textit{\colorcell{81.52}{80.86}} &  & \textcolor{atomictangerine!60}{$\bullet$} \textit{\colorcell{60.63}{58.53}} &  & \textcolor{white}{$\bullet$} \textit{\colorcell{90.83}{91.91}} & \textcolor{white}{$\bullet$} \textit{\colorcell{79.37}{81.17}} &  & \textcolor{atomictangerine!60}{$\bullet$} \textit{\colorcell{51.11}{49.49}} \\
\bottomrule
\end{tabular}}
\caption{Impact of the systems used for candidate generation on WMT'22 performance in the multi-system setting after undergoing SFT and CPO optimization. Values in \textit{italic} font denote statistically significant differences between all-systems-based alignment and alignment with one system removed, at the $5\%$ significance level, based on one-tailed paired Student's $t$-tests. Evaluation metrics and color codes are the same as in Table~\ref{tab:results}.}
\label{tab:no_sys_xps}
\end{table*}

\paragraph{Neural-based alignment improves downstream performance.}
Table~\ref{tab:results} shows that when aligning with neural metrics (xCOMET-QE or CometKiwi), both SFT on preferred translations and CPO consistently improve performance on the alignment metric across language pairs. We also observe that aligning on xCOMET-QE improves results on CometKiwi, and vice-versa. We hypothesize this may be the result of high correlation between different neural metrics, as they are typically trained on similar data. Overall, these results demonstrate that alignment-based techniques can achieve similar objectives to those of quality-aware decoding approaches like MBR, even though the target metric is only indirectly optimized.

\paragraph{CPO induces adverse metric effects.}
In Table~\ref{tab:results}, we observe that when aligning with neural metrics, CPO yields significantly greater improvements on the alignment metric compared to SFT. The inclusion of the reject option seems to offer additional benefits over the traditional SFT objective in this context. However, aligning with CPO also introduces adverse effects between neural and lexical metrics for out-of-English translations. More specifically, and consistent with the findings of \citet{xu2024contrastive}, aligning on neural metrics negatively impacts lexical metrics. Importantly, this is further evidence to support recommendations provided in \cite{kocmi2024navigating}: even though, in most cases, neural and lexical MT evaluation metrics should be positively correlated, we should employ caution when using the same metric for evaluation that was used during training/inference. Nevertheless and perhaps more interestingly, it turns out SFT does not produce such effects, raising the question of whether these contradictory evaluation dynamics seen with CPO stem from the learning objective itself or the mix of candidate systems used.

\paragraph{Lexical alignment fails to improve downstream performance.}
Table~\ref{tab:results} shows that preference-based lexical alignment\footnote{When performing alignment using a lexical metric like chrF, the chosen translation is by definition the gold reference as long as it is present in the pool of candidates. The translation with the lowest chrF score among the remaining systems is then rejected.} behaves differently compared to neural alignment. Specifically, SFT results are roughly stagnant, showing a slight decrease in chrF for into-English translations and a slight increase for out-of-English translations. In contrast, CPO results in a steep drop across the metric board for both into- and out-of-English translations. Using the gold reference as the chosen system appears to impair downstream performance, especially when performing alignment using CPO.

\subsection{Impact of the Candidate Systems}
\label{sec:cand_sys_impact}

\begin{table*}[t]
\renewcommand\arraystretch{0.9}
\centering
\small
\resizebox{0.99\textwidth}{!}
{\begin{tabular}{lccccccccc}
\toprule
 & \multicolumn{4}{c}{\texttt{xx-en}} &  & \multicolumn{4}{c}{\texttt{en-xx}} \\
\noalign{\vskip 0.5pt}
\cline{2-5}
\cline{7-10}\noalign{\vskip 2pt}
 & \multicolumn{2}{c}{Neural} &  & Lexical &  & \multicolumn{2}{c}{Neural} &  & Lexical \\
\noalign{\vskip 0.5pt}
\cline{2-3}
\cline{5-5}
\cline{7-8}
\cline{10-10}\noalign{\vskip 2pt}
 & xCOMET-QE & CometKiwi &  & chrF &  & xCOMET-QE & CometKiwi &  & chrF \\
\midrule
\textbf{Base} & \textcolor{white}{$\bullet$} \colorcell{87.80}{87.80} & \textcolor{white}{$\bullet$} \colorcell{80.86}{80.86} &  & \textcolor{white}{$\bullet$} \colorcell{58.53}{58.53} &  & \textcolor{white}{$\bullet$} \colorcell{91.91}{91.91} & \textcolor{white}{$\bullet$} \colorcell{81.17}{81.17} &  & \textcolor{white}{$\bullet$} \colorcell{49.49}{49.49} \\\midrule
\multicolumn{10}{c}{\textit{Chosen system set to} \textbf{Base}} \\\cdashlinelr{1-10}
SFT & \textcolor{atomictangerine!60}{$\bullet$} \colorcell{88.17}{87.80} & \textcolor{white}{$\bullet$} \colorcell{81.08}{80.86} &  & \textcolor{white}{$\bullet$} \colorcell{58.91}{58.53} &  & \textcolor{atomictangerine!60}{$\bullet$} \colorcell{91.94}{91.91} & \textcolor{white}{$\bullet$} \colorcell{81.21}{81.17} &  & \textcolor{white}{$\bullet$} \colorcell{49.35}{49.49} \\
CPO & \textcolor{atomictangerine!60}{$\bullet$} \textit{\colorcell{87.94}{87.80}} & \textcolor{white}{$\bullet$} \colorcell{81.02}{80.86} &  & \textcolor{white}{$\bullet$} \textit{\colorcell{58.62}{58.53}} &  & \textcolor{atomictangerine!60}{$\bullet$} \textit{\colorcell{91.75}{91.91}} & \textcolor{white}{$\bullet$} \textit{\colorcell{81.06}{81.17}} &  & \textcolor{white}{$\bullet$} \textit{\colorcell{48.56}{49.49}} \\\midrule
\multicolumn{10}{c}{\textit{Chosen system set to} \textbf{Ref}} \\\cdashlinelr{1-10}
SFT & \textcolor{atomictangerine!60}{$\bullet$} \colorcell{88.04}{87.80} & \textcolor{white}{$\bullet$} \colorcell{81.06}{80.86} &  & \textcolor{white}{$\bullet$} \colorcell{57.73}{58.53} &  & \textcolor{atomictangerine!60}{$\bullet$} \colorcell{92.35}{91.91} & \textcolor{white}{$\bullet$} \colorcell{81.94}{81.17} &  & \textcolor{white}{$\bullet$} \colorcell{50.12}{49.49} \\
CPO & \textcolor{atomictangerine!60}{$\bullet$} \textit{\colorcell{81.95}{87.80}} & \textcolor{white}{$\bullet$} \textit{\colorcell{77.86}{80.86}} &  & \textcolor{white}{$\bullet$} \textit{\colorcell{48.75}{58.53}} &  & \textcolor{atomictangerine!60}{$\bullet$} \textit{\colorcell{86.97}{91.91}} & \textcolor{white}{$\bullet$} \textit{\colorcell{80.01}{81.17}} &  & \textcolor{white}{$\bullet$} \textit{\colorcell{39.81}{49.49}} \\\midrule
\multicolumn{10}{c}{\textit{Chosen system set to} \textbf{GPT-4}} \\\cdashlinelr{1-10}
SFT & \textcolor{atomictangerine!60}{$\bullet$} \colorcell{89.81}{87.80} & \textcolor{white}{$\bullet$} \colorcell{81.67}{80.86} &  & \textcolor{white}{$\bullet$} \colorcell{60.53}{58.53} &  & \textcolor{atomictangerine!60}{$\bullet$} \colorcell{91.96}{91.91} & \textcolor{white}{$\bullet$} \colorcell{80.83}{81.17} &  & \textcolor{white}{$\bullet$} \colorcell{50.73}{49.49} \\
CPO & \textcolor{atomictangerine!60}{$\bullet$} \colorcell{89.69}{87.80} & \textcolor{white}{$\bullet$} \textit{\colorcell{80.99}{80.86}} &  & \textcolor{white}{$\bullet$} \colorcell{60.42}{58.53} &  & \textcolor{atomictangerine!60}{$\bullet$} \textit{\colorcell{90.50}{91.91}} & \textcolor{white}{$\bullet$} \textit{\colorcell{78.81}{81.17}} &  & \textcolor{white}{$\bullet$} \textit{\colorcell{50.22}{49.49}} \\
\bottomrule
\end{tabular}}
\caption{Impact of imposing the chosen system on WMT'22 downstream performance in the multi-system setting. Values in \textit{italic} font denote statistically significant differences between SFT- and CPO-based alignment at the $5\%$ significance level, based on one-tailed paired Student's $t$-tests. Evaluation metrics and color codes are the same as in Table~\ref{tab:results}.}
\label{tab:chosen_sys_impact}
\end{table*}

We now turn to investigating how much the success of alignment-based fine-tuning depends on the choice of the candidate systems. Unless otherwise specified, we use xCOMET-QE as the alignment metric and examine the performance impact of withdrawing systems from the candidate pool. We perform SFT and CPO on the newly created datasets. We report results in Table~\ref{tab:no_sys_xps}.

\paragraph{The choice of the candidate systems impacts alignment performance.}
Table~\ref{tab:no_sys_xps} shows that for both SFT- and CPO-based methods, removing systems from the pool of candidates significantly affects performance on the alignment metric.  This is particularly the case for out-of-English translation with CPO optimization. Notably, removing GPT-4 has the strongest negative impact on downstream xCOMET-QE. This is expected as it is the highest-quality system among the system candidates~(see Table~\ref{tab:data_stats} in Appendix~\ref{apdx:data_details}).

\paragraph{Some candidate systems can be harmful to preference-based alignment.} 
In Section~\ref{sec:top_level}, we observed CPO negatively impacts \texttt{en-xx} chrF when aligning on neural metrics, unlike SFT on preferred translations. Table~\ref{tab:no_sys_xps} suggests this may stem from including gold references in the candidate system pool: removing them eliminates this adverse effect. We also noted in Section~\ref{sec:top_level} that lexical alignment fails to improve downstream chrF, with sharp decreases with CPO. This issue is resolved by removing gold references. Overall, candidate system choice affects alignment effectiveness and downstream metric consistency, with CPO showing higher sensitivity to preference settings than SFT.

\subsection{Impact of the Chosen System}
\label{sec:chos_sys_impact}

To complement findings from Section~\ref{sec:cand_sys_impact} and further characterize the sensitivity of preference-based alignment, we propose examining downstream performance when the chosen system is fixed to a single system. We create three preference datasets based on xCOMET-QE, in which we either impose the base model, reference or GPT-4 as the chosen system. When applicable, the rejected translation is selected from the remaining systems~(if one has a lower xCOMET-QE than the chosen system); otherwise, the sample is discarded.

\paragraph{CPO is not robust to the preference setting.}
In contrast to the observations made in Section~\ref{sec:top_level}, Table~\ref{tab:chosen_sys_impact} shows that, under this setup, CPO fails to outperform SFT for both \texttt{xx-en} and \texttt{en-xx} translations. When systematically choosing base translations, CPO is unable to surpass the trivial SFT setting where the base model is fine-tuned on its own translations.\footnote{As expected, performing SFT on a model's own greedy predictions has minimal impact on downstream performance.} Moreove, downstream CPO performance significantly declines when gold references are chosen, underperforming the non-aligned model across all metrics, even including the alignment metric. These results reinforce the claims made in Section~\ref{sec:cand_sys_impact} and indicate a lack of robustness of CPO compared to SFT. In the following section (Section~\ref{sec:mono_system}), we demonstrate that this instability observed with CPO can be mitigated by using a more normalized preference setting, relying only on the base model for candidate generation.

\section{Mono-System Preference Fine-Tuning}
\label{sec:mono_system}

\begin{table*}[t]
\renewcommand\arraystretch{0.9}
\small
\centering
\resizebox{0.99\textwidth}{!}
{\begin{tabular}{lccccccccc}
\toprule
 & \multicolumn{4}{c}{\texttt{xx-en}} &  & \multicolumn{4}{c}{\texttt{en-xx}} \\
\noalign{\vskip 0.5pt}
\cline{2-5}
\cline{7-10}
\noalign{\vskip 2pt}
 & \multicolumn{2}{c}{Neural} &  & Lexical &  & \multicolumn{2}{c}{Neural} &  & Lexical \\
\noalign{\vskip 0.5pt}
\cline{2-3}
\cline{5-5}
\cline{7-8}
\cline{10-10}
\noalign{\vskip 2pt}
 & xCOMET-QE & CometKiwi &  & chrF &  & xCOMET-QE & CometKiwi &  & chrF \\
\midrule
\textbf{Base} & \textcolor{white}{$\bullet$} \colorcell{87.80}{87.80} & \textcolor{white}{$\bullet$} \colorcell{80.86}{80.86} &  & \textcolor{white}{$\bullet$} \colorcell{58.53}{58.53} &  & \textcolor{white}{$\bullet$} \colorcell{91.91}{91.91} & \textcolor{white}{$\bullet$} \colorcell{81.17}{81.17} &  & \textcolor{white}{$\bullet$} \colorcell{49.49}{49.49} \\\midrule
\multicolumn{10}{c}{\textit{Optimization via} \textbf{SFT}} \\\cdashlinelr{1-10}
Multi-system & \textcolor{atomictangerine!60}{$\bullet$} \colorcell{89.13}{87.80} & \textcolor{white}{$\bullet$} \colorcell{81.49}{80.86} &  & \textcolor{white}{$\bullet$} \colorcell{59.82}{58.53} &  & \textcolor{atomictangerine!60}{$\bullet$} \colorcell{92.38}{91.91} & \textcolor{white}{$\bullet$} \colorcell{81.67}{81.17} &  & \textcolor{white}{$\bullet$} \colorcell{50.28}{49.49} \\
Mono-system & \textcolor{atomictangerine!60}{$\bullet$} \textit{\colorcell{88.51}{87.80}} & \textcolor{white}{$\bullet$} \textit{\colorcell{81.29}{80.86}} &  & \textcolor{white}{$\bullet$} \textit{\colorcell{59.05}{58.53}} &  & \textcolor{atomictangerine!60}{$\bullet$} \textit{\colorcell{92.17}{91.91}} & \textcolor{white}{$\bullet$} \colorcell{81.54}{81.17} &  & \textcolor{white}{$\bullet$} \textit{\colorcell{49.41}{49.49}} \\\midrule
\multicolumn{10}{c}{\textit{Optimization via} \textbf{CPO}} \\\cdashlinelr{1-10}
Multi-system & \textcolor{atomictangerine!60}{$\bullet$} \colorcell{89.95}{87.80} & \textcolor{white}{$\bullet$} \colorcell{81.89}{80.86} &  & \textcolor{white}{$\bullet$} \colorcell{59.83}{58.53} &  & \textcolor{atomictangerine!60}{$\bullet$} \colorcell{92.75}{91.91} & \textcolor{white}{$\bullet$} \colorcell{83.60}{81.17} &  & \textcolor{white}{$\bullet$} \colorcell{47.69}{49.49} \\
Mono-system & \textcolor{atomictangerine!60}{$\bullet$} \textit{\colorcell{89.35}{87.80}} & \textcolor{white}{$\bullet$} \colorcell{81.80}{80.86} &  & \textcolor{white}{$\bullet$} \textit{\colorcell{59.52}{58.53}} &  & \textcolor{atomictangerine!60}{$\bullet$} \colorcell{92.69}{91.91} & \textcolor{white}{$\bullet$} \textit{\colorcell{82.91}{81.17}} &  & \textcolor{white}{$\bullet$} \textit{\colorcell{49.02}{49.49}} \\
Mono-system (opt.) & \textcolor{atomictangerine!60}{$\bullet$} \textit{\colorcell{89.58}{87.80}} & \textcolor{white}{$\bullet$} \colorcell{81.97}{80.86} &  & \textcolor{white}{$\bullet$} \textit{\colorcell{59.65}{58.53}} &  & \textcolor{atomictangerine!60}{$\bullet$} \colorcell{92.87}{91.91} & \textcolor{white}{$\bullet$} \colorcell{83.47}{81.17} &  & \textcolor{white}{$\bullet$} \textit{\colorcell{49.11}{49.49}} \\
\bottomrule
\end{tabular}}
\caption{Comparison between multi- and mono-system fine-tuning on WMT'22 test data. Alignment is performed on xCOMET-QE for both SFT and CPO. Mono-system (opt.) denotes the model fine-tuned on optimized mono-system preference data. Values in \textit{italic} font denote statistically significant differences between multi-system- and mono-system-based alignment at the $5\%$ significance level. Evaluation metrics and color codes are the same as in Table~\ref{tab:results}, based on one-tailed paired Student's $t$-tests.}
\label{tab:mono_vs_multi}
\end{table*}

So far, we have exclusively focused on multi-system alignment, which involves using external models for candidate generation and preference dataset building. Although this approach is common for metric alignment \cite{luong-manning-2015-stanford, sennrich-etal-2016-improving, xu2024contrastive}, some works have shown that a model can be aligned effectively using only its own outputs \cite{yang2023direct, yuan2024selfrewardinglanguagemodels, dubey2024llama3herdmodels}. In this section, we propose to take a closer look at this strategy and identify its potential advantages and disadvantages compared to the multi-system approach. We use xCOMET-QE as the alignment metric. To ensure a fair comparison, we first generate the mono-system dataset to approximately replicate the properties of the multi-system dataset regarding the alignment metric.\footnote{The created mono-system dataset has an average rejected/chosen xCOMET-QE of $87.8/97.3$, compared to $87.9/97.2$ for the multi-system dataset (Table~\ref{tab:data_stats}).} Details on the construction of mono-system preference datasets are given in Section~\ref{sec:xp_setup} and Appendix \ref{apdx:pref_build}.

\subsection{Comparison With Multi-System Alignment}
\label{sec:mono_multi}

\paragraph{Mono-system alignment improves downstream performance.}
Table~\ref{tab:mono_vs_multi} shows that performing SFT and CPO on a mono-system dataset using xCOMET-QE for alignment results in improved downstream performance across all neural metrics compared to the base model, as observed in the multi-system scenario (Section~\ref{sec:top_level}). This finding highlights the effectiveness of alignment techniques even when using only the model's own translations for candidate generation, without needing access to high-quality external systems. This is particularly relevant in practical scenarios in which such access may be limited or unavailable.

\paragraph{CPO consistently outperforms SFT on neural metrics.}
Similar to when relying on multiple systems for candidate generation, we observe in Table~\ref{tab:mono_vs_multi} that CPO outperforms SFT regarding downstream performance on neural metrics. This finding reinforces the observation made in Section~\ref{sec:top_level} and tends to confirm the superiority of the CPO objective over SFT on preferred translations in optimizing neural-based alignment performance. 

\begin{figure*}[t]
  \centering
  \includegraphics[width=0.93\textwidth]{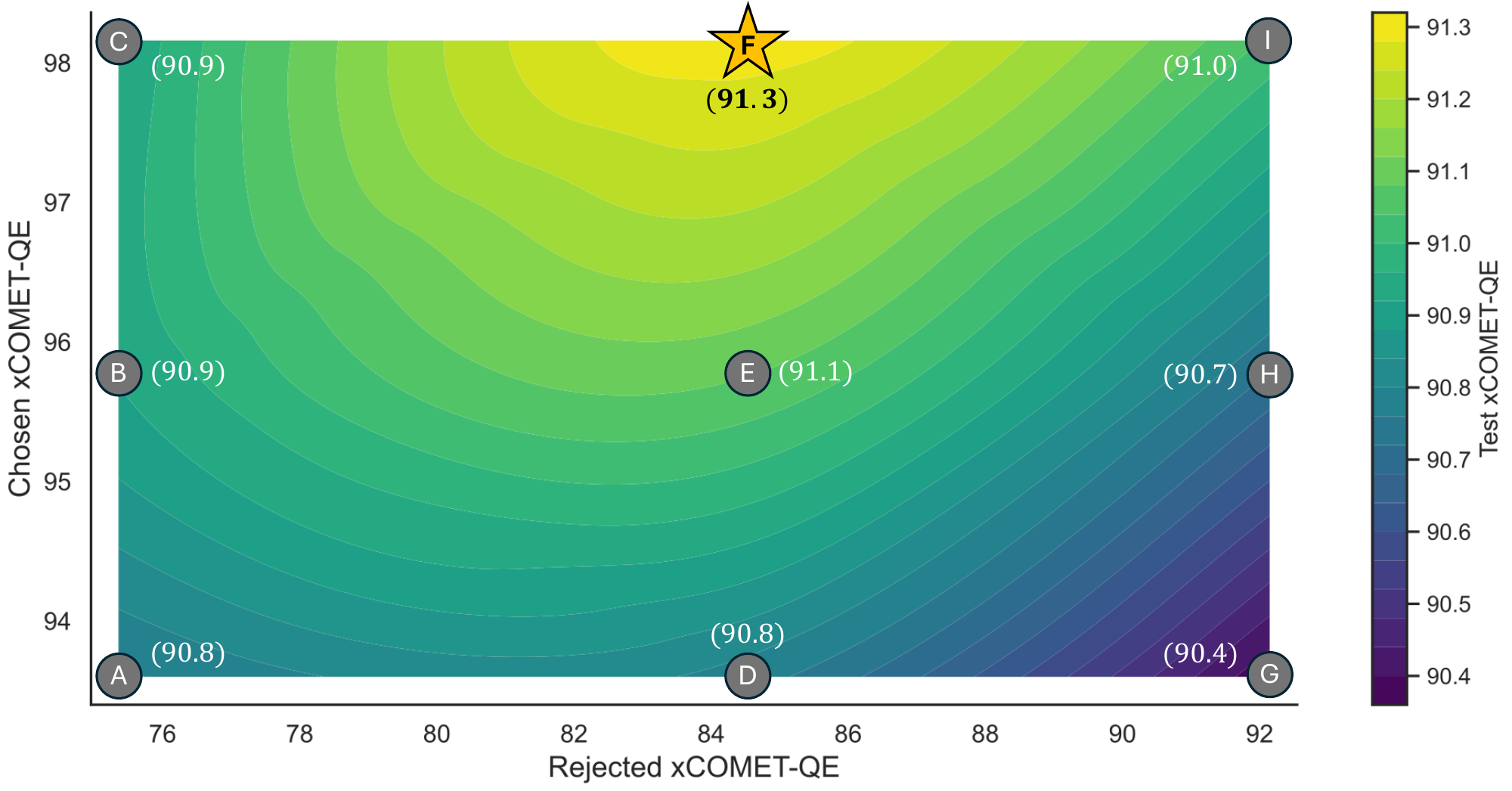}
  \caption{Impact of chosen and rejected option quality on downstream performance, using xCOMET-QE for alignment and evaluation. The chart is derived by linearly interpolating results from nine preference datasets (points A to I), each with different average rejected and chosen qualities. Test performance on WMT'22 (average across all language pairs) is reported in brackets. Example: point C (avg. rejected xCOMET-QE: $75.4$, avg. chosen: $98.2$) achieves $90.9$ xCOMET-QE on WMT'22 test data.}
  \label{fig:qual_study}
\end{figure*}

\paragraph{Mono-system alignment slightly underperforms multi-system alignment.} Table~\ref{tab:mono_vs_multi} shows that while mono-system alignment increases downstream performance on neural metrics, the improvement levels are not as high as in the multi-system setting. Despite the mono- and multi-system preference datasets being built with the same alignment metric properties, having translations from different distributions, particularly from GPT-4 (cf. Section~\ref{sec:cand_sys_impact} and Table~\ref{tab:no_sys_xps}), appears to add value for achieving optimized alignment effectiveness.

\paragraph{Removing external systems almost eliminates the adverse metric effects observed with CPO.}
In Section~\ref{sec:top_level}, we showed that multi-system neural alignment using CPO greatly impacts lexical performance for out-of-English translations. Table~\ref{tab:mono_vs_multi} demonstrates that mono-system alignment almost completely mitigates these negative effects. While there is still a slight decrease in \texttt{en-xx} chrF, it is much smaller compared to the multi-system scenario. This confirms the findings from Sections \ref{sec:cand_sys_impact} and \ref{sec:chos_sys_impact} that CPO is sensitive to the preference setting, but also shows that relying solely on candidate translations from the base model limits adverse effects on downstream metric consistency. A possible explanation is that candidate translations from the same system distribution tend to have similar properties, thereby reducing the likelihood of observing high lexical instability when performing alignment based on a neural metric like xCOMET-QE.

\paragraph{The mono-system approach offers better control over the alignment process.}
Specifically, mono-system alignment provides more fine-grained control over the respective qualities of the chosen and rejected options. This setting allows for tuning these qualities to maximize post-alignment performance, which is not possible when using a limited number of external systems. This aspect is further explored in the following section (Section~\ref{sec:qual_study}).

\subsection{Optimizing the Preference Data}
\label{sec:qual_study}

In this final experiment, we examine how the quality of chosen and rejected options affects downstream performance. We build nine preference datasets, each with varying average xCOMET-QE scores for chosen and rejected options. The hypotheses' average qualities are categorized into three groups: High, Mid, and Low. As detailed in Section~\ref{sec:pref_data}, the quality of the chosen (resp. rejected) option is always ensured to be above (resp. below) the quality of the base translation. The statistics of the created datasets are summarized in Appendix \ref{apdx:pref_build} (Table~\ref{tab:data_stats}).

\paragraph{The respective qualities of the rejected and chosen options have a significant impact on post-CPO performance.}
Figure~\ref{fig:qual_study} highlights the need to closely monitor the qualities of chosen and rejected options to fully leverage the mono-system approach. Specifically, several properties of preference data were found to negatively impact post-CPO performance: (i) a chosen option of too low quality, (ii) an extremely low or high quality of the rejected option, and (iii) too wide a gap between the qualities of the rejected and chosen options.

\paragraph{Optimizing preference data yields competitive performance to multi-system setting.}
Figure~\ref{fig:qual_study} shows that for effective metric alignment with CPO, the rejected option's quality should be moderate (neither too high nor too low), while the chosen option's quality should be as high as possible. Specifically, optimal test performance was obtained with rejected options average around 90\% ($\Delta=-10\%$) of the base model's quality, and chosen options averaging around 105\% ($\Delta=+5\%$). Under this scenario, we show that performance levels can match those in the multi-system setting while maintaining consistency with lexical scores  (Table~\ref{tab:mono_vs_multi}). However, these results also highlight the complexity of achieving optimal preference-based alignment and get the most of the reject option.

\section{Conclusion}

Our experiments revealed several key findings. Firstly, we showed that preference-based alignment, specifically using CPO, globally outperforms SFT on high-quality data in terms of improving neural evaluation metrics. However, we identified significant drawbacks when relying on multiple systems for preference data generation, revealing adverse effects between neural and lexical metrics, and highlighting a lack of robustness in preference-based alignment compared to the SFT approach. Finally, we showed that using candidate translations all originating from the same system distribution, specifically the base model, can be an effective strategy for gaining more control over preference-based fine-tuning. This approach achieves performance comparable to using multiple external systems while ensuring better consistency across evaluation metrics. In a nutshell, while preference-based alignment techniques hold promise for improving MT quality, careful consideration must be given to the choice of candidate translations, the learning objective, and the potential trade-offs regarding downstream metric consistency.

\section*{Limitations}

In this work, we conducted extensive experiments to assess the impact of preference-based fine-tuning on downstream translation quality. For efficiency and practicality, we focused on the experimental setup detailed by \citet{xu2024contrastive}, which utilizes three systems for candidate generation. Similarly, we used the same evaluation metrics and datasets. Future experiments could benefit from validating our findings using different model families, a broader range of alignment and evaluation metrics, and additional translation datasets, for instance including other languages.

Additionally, in the mono-system setting, we explored the impact of varying the qualities of chosen and rejected options and derived general insights on optimizing preference data. Further research could involve using different datasets, models, and alignment metrics to characterize more precisely the factors that influence downstream performance in this specific scenario. This approach could lead to a deeper mathematical understanding of the elements that affect performance in preference-based fine-tuning, resulting in more robust and scalable optimization techniques.

Finally, our evaluation relied on automatic metrics, both lexical and neural, with the latter closely approximating human judgments but still being unable to fully replace them. Given their imperfect correlation with human preferences, future work could benefit from additional human evaluation of outputs obtained via the approaches we studied to get an even deeper understanding of post-alignment downstream performance dynamics.

\section*{Ethics Statement}

Our work aims to investigate the mechanisms of model alignment to enhance transparency in the field of automatic translation. We believe this effort improves the interpretability of model outputs, which is beneficial for ethical considerations. Additionally, our analysis is distinctly multilingual, with an emphasis on low-resource languages, contributing to expanding the scope of MT. We have identified no potential negative societal impacts from our work.

\section*{Acknowledgements}

Training compute was provided by the Jean Zay supercomputer, operated by GENCI IDRIS, through compute grants 2023-AD011014668R1 and AD010614770, as well as by the Adastra supercomputer through projects c1615122, cad15031, and cad14770. Part of this work was also supported by the EU's Horizon Europe Research and Innovation Actions (UTTER, contract 101070631), by the DECOLLAGE project (ERC-2022-CoG 101088763), and by the Portuguese Recovery and Resilience Plan through project C645008882-00000055 (Center for Responsible AI).

\bibliography{main}

\begin{thebibliography}{47}
\expandafter\ifx\csname natexlab\endcsname\relax\def\natexlab#1{#1}\fi

\bibitem[{Alves et~al.(2023)Alves, Guerreiro, Alves, Pombal, Rei, de~Souza, Colombo, and Martins}]{alves2023steering}
Duarte~M Alves, Nuno~M Guerreiro, Jo{\~a}o Alves, Jos{\'e} Pombal, Ricardo Rei, Jos{\'e}~GC de~Souza, Pierre Colombo, and Andr{\'e}~FT Martins. 2023.
\newblock Steering large language models for machine translation with finetuning and in-context learning.
\newblock \emph{arXiv preprint arXiv:2310.13448}.

\bibitem[{Alves et~al.(2024)Alves, Pombal, Guerreiro, Martins, Alves, Farajian, Peters, Rei, Fernandes, Agrawal et~al.}]{alves2024tower}
Duarte~M Alves, Jos{\'e} Pombal, Nuno~M Guerreiro, Pedro~H Martins, Jo{\~a}o Alves, Amin Farajian, Ben Peters, Ricardo Rei, Patrick Fernandes, Sweta Agrawal, et~al. 2024.
\newblock Tower: An open multilingual large language model for translation-related tasks.
\newblock \emph{arXiv preprint arXiv:2402.17733}.

\bibitem[{Amrhein and Sennrich(2022)}]{amrhein2022identifying}
Chantal Amrhein and Rico Sennrich. 2022.
\newblock Identifying weaknesses in machine translation metrics through minimum bayes risk decoding: A case study for comet.
\newblock \emph{arXiv preprint arXiv:2202.05148}.

\bibitem[{Banerjee and Lavie(2005)}]{banerjee-lavie-2005-meteor}
Satanjeev Banerjee and Alon Lavie. 2005.
\newblock \href {https://aclanthology.org/W05-0909} {{METEOR}: An automatic metric for {MT} evaluation with improved correlation with human judgments}.
\newblock In \emph{Proceedings of the {ACL} Workshop on Intrinsic and Extrinsic Evaluation Measures for Machine Translation and/or Summarization}, pages 65--72, Ann Arbor, Michigan. Association for Computational Linguistics.

\bibitem[{Briakou et~al.(2024)Briakou, Luo, Cherry, and Freitag}]{briakou2024translating}
Eleftheria Briakou, Jiaming Luo, Colin Cherry, and Markus Freitag. 2024.
\newblock Translating step-by-step: Decomposing the translation process for improved translation quality of long-form texts.
\newblock \emph{arXiv preprint arXiv:2409.06790}.

\bibitem[{Dubey et~al.(2024)Dubey, Jauhri, Pandey, Kadian, Al-Dahle, Letman, Mathur, Schelten, Yang, Fan, Goyal, Hartshorn, Yang, Mitra, Sravankumar, Korenev, Hinsvark, Rao, Zhang, Rodriguez, Gregerson, Spataru, Roziere, Biron, Tang, Chern, Caucheteux, Nayak, Bi, Marra, McConnell, Keller, Touret, Wu, Wong, Ferrer, Nikolaidis, Allonsius, Song, Pintz, Livshits, Esiobu, Choudhary, Mahajan, Garcia-Olano, Perino, Hupkes, Lakomkin, AlBadawy, Lobanova, Dinan, Smith, Radenovic, Zhang, Synnaeve, Lee, Anderson, Nail, Mialon, Pang, Cucurell, Nguyen, Korevaar, Xu, Touvron, Zarov, Ibarra, Kloumann, Misra, Evtimov, Copet, Lee, Geffert, Vranes, Park, Mahadeokar, Shah, van~der Linde, Billock, Hong, Lee, Fu, Chi, Huang, Liu, Wang, Yu, Bitton, Spisak, Park, Rocca, Johnstun, Saxe, Jia, Alwala, Upasani, Plawiak, Li, Heafield, Stone, El-Arini, Iyer, Malik, Chiu, Bhalla, Rantala-Yeary, van~der Maaten, Chen, Tan, Jenkins, Martin, Madaan, Malo, Blecher, Landzaat, de~Oliveira, Muzzi, Pasupuleti, Singh, Paluri, Kardas, Oldham, Rita,
  Pavlova, Kambadur, Lewis, Si, Singh, Hassan, Goyal, Torabi, Bashlykov, Bogoychev, Chatterji, Duchenne, Çelebi, Alrassy, Zhang, Li, Vasic, Weng, Bhargava, Dubal, Krishnan, Koura, Xu, He, Dong, Srinivasan, Ganapathy, Calderer, Cabral, Stojnic, Raileanu, Girdhar, Patel, Sauvestre, Polidoro, Sumbaly, Taylor, Silva, Hou, Wang, Hosseini, Chennabasappa, Singh, Bell, Kim, Edunov, Nie, Narang, Raparthy, Shen, Wan, Bhosale, Zhang, Vandenhende, Batra, Whitman, Sootla, Collot, Gururangan, Borodinsky, Herman, Fowler, Sheasha, Georgiou, Scialom, Speckbacher, Mihaylov, Xiao, Karn, Goswami, Gupta, Ramanathan, Kerkez, Gonguet, Do, Vogeti, Petrovic, Chu, Xiong, Fu, Meers, Martinet, Wang, Tan, Xie, Jia, Wang, Goldschlag, Gaur, Babaei, Wen, Song, Zhang, Li, Mao, Coudert, Yan, Chen, Papakipos, Singh, Grattafiori, Jain, Kelsey, Shajnfeld, Gangidi, Victoria, Goldstand, Menon, Sharma, Boesenberg, Vaughan, Baevski, Feinstein, Kallet, Sangani, Yunus, Lupu, Alvarado, Caples, Gu, Ho, Poulton, Ryan, Ramchandani, Franco, Saraf,
  Chowdhury, Gabriel, Bharambe, Eisenman, Yazdan, James, Maurer, Leonhardi, Huang, Loyd, Paola, Paranjape, Liu, Wu, Ni, Hancock, Wasti, Spence, Stojkovic, Gamido, Montalvo, Parker, Burton, Mejia, Wang, Kim, Zhou, Hu, Chu, Cai, Tindal, Feichtenhofer, Civin, Beaty, Kreymer, Li, Wyatt, Adkins, Xu, Testuggine, David, Parikh, Liskovich, Foss, Wang, Le, Holland, Dowling, Jamil, Montgomery, Presani, Hahn, Wood, Brinkman, Arcaute, Dunbar, Smothers, Sun, Kreuk, Tian, Ozgenel, Caggioni, Guzmán, Kanayet, Seide, Florez, Schwarz, Badeer, Swee, Halpern, Thattai, Herman, Sizov, Guangyi, Zhang, Lakshminarayanan, Shojanazeri, Zou, Wang, Zha, Habeeb, Rudolph, Suk, Aspegren, Goldman, Damlaj, Molybog, Tufanov, Veliche, Gat, Weissman, Geboski, Kohli, Asher, Gaya, Marcus, Tang, Chan, Zhen, Reizenstein, Teboul, Zhong, Jin, Yang, Cummings, Carvill, Shepard, McPhie, Torres, Ginsburg, Wang, Wu, U, Saxena, Prasad, Khandelwal, Zand, Matosich, Veeraraghavan, Michelena, Li, Huang, Chawla, Lakhotia, Huang, Chen, Garg, A, Silva, Bell,
  Zhang, Guo, Yu, Moshkovich, Wehrstedt, Khabsa, Avalani, Bhatt, Tsimpoukelli, Mankus, Hasson, Lennie, Reso, Groshev, Naumov, Lathi, Keneally, Seltzer, Valko, Restrepo, Patel, Vyatskov, Samvelyan, Clark, Macey, Wang, Hermoso, Metanat, Rastegari, Bansal, Santhanam, Parks, White, Bawa, Singhal, Egebo, Usunier, Laptev, Dong, Zhang, Cheng, Chernoguz, Hart, Salpekar, Kalinli, Kent, Parekh, Saab, Balaji, Rittner, Bontrager, Roux, Dollar, Zvyagina, Ratanchandani, Yuvraj, Liang, Alao, Rodriguez, Ayub, Murthy, Nayani, Mitra, Li, Hogan, Battey, Wang, Maheswari, Howes, Rinott, Bondu, Datta, Chugh, Hunt, Dhillon, Sidorov, Pan, Verma, Yamamoto, Ramaswamy, Lindsay, Lindsay, Feng, Lin, Zha, Shankar, Zhang, Zhang, Wang, Agarwal, Sajuyigbe, Chintala, Max, Chen, Kehoe, Satterfield, Govindaprasad, Gupta, Cho, Virk, Subramanian, Choudhury, Goldman, Remez, Glaser, Best, Kohler, Robinson, Li, Zhang, Matthews, Chou, Shaked, Vontimitta, Ajayi, Montanez, Mohan, Kumar, Mangla, Albiero, Ionescu, Poenaru, Mihailescu, Ivanov, Li, Wang,
  Jiang, Bouaziz, Constable, Tang, Wang, Wu, Wang, Xia, Wu, Gao, Chen, Hu, Jia, Qi, Li, Zhang, Zhang, Adi, Nam, Yu, Wang, Hao, Qian, He, Rait, DeVito, Rosnbrick, Wen, Yang, and Zhao}]{dubey2024llama3herdmodels}
Abhimanyu Dubey, Abhinav Jauhri, Abhinav Pandey, Abhishek Kadian, Ahmad Al-Dahle, Aiesha Letman, Akhil Mathur, Alan Schelten, Amy Yang, Angela Fan, Anirudh Goyal, Anthony Hartshorn, Aobo Yang, Archi Mitra, Archie Sravankumar, Artem Korenev, Arthur Hinsvark, Arun Rao, Aston Zhang, Aurelien Rodriguez, Austen Gregerson, Ava Spataru, Baptiste Roziere, Bethany Biron, Binh Tang, Bobbie Chern, Charlotte Caucheteux, Chaya Nayak, Chloe Bi, Chris Marra, Chris McConnell, Christian Keller, Christophe Touret, Chunyang Wu, Corinne Wong, Cristian~Canton Ferrer, Cyrus Nikolaidis, Damien Allonsius, Daniel Song, Danielle Pintz, Danny Livshits, David Esiobu, Dhruv Choudhary, Dhruv Mahajan, Diego Garcia-Olano, Diego Perino, Dieuwke Hupkes, Egor Lakomkin, Ehab AlBadawy, Elina Lobanova, Emily Dinan, Eric~Michael Smith, Filip Radenovic, Frank Zhang, Gabriel Synnaeve, Gabrielle Lee, Georgia~Lewis Anderson, Graeme Nail, Gregoire Mialon, Guan Pang, Guillem Cucurell, Hailey Nguyen, Hannah Korevaar, Hu~Xu, Hugo Touvron, Iliyan Zarov,
  Imanol~Arrieta Ibarra, Isabel Kloumann, Ishan Misra, Ivan Evtimov, Jade Copet, Jaewon Lee, Jan Geffert, Jana Vranes, Jason Park, Jay Mahadeokar, Jeet Shah, Jelmer van~der Linde, Jennifer Billock, Jenny Hong, Jenya Lee, Jeremy Fu, Jianfeng Chi, Jianyu Huang, Jiawen Liu, Jie Wang, Jiecao Yu, Joanna Bitton, Joe Spisak, Jongsoo Park, Joseph Rocca, Joshua Johnstun, Joshua Saxe, Junteng Jia, Kalyan~Vasuden Alwala, Kartikeya Upasani, Kate Plawiak, Ke~Li, Kenneth Heafield, Kevin Stone, Khalid El-Arini, Krithika Iyer, Kshitiz Malik, Kuenley Chiu, Kunal Bhalla, Lauren Rantala-Yeary, Laurens van~der Maaten, Lawrence Chen, Liang Tan, Liz Jenkins, Louis Martin, Lovish Madaan, Lubo Malo, Lukas Blecher, Lukas Landzaat, Luke de~Oliveira, Madeline Muzzi, Mahesh Pasupuleti, Mannat Singh, Manohar Paluri, Marcin Kardas, Mathew Oldham, Mathieu Rita, Maya Pavlova, Melanie Kambadur, Mike Lewis, Min Si, Mitesh~Kumar Singh, Mona Hassan, Naman Goyal, Narjes Torabi, Nikolay Bashlykov, Nikolay Bogoychev, Niladri Chatterji, Olivier
  Duchenne, Onur Çelebi, Patrick Alrassy, Pengchuan Zhang, Pengwei Li, Petar Vasic, Peter Weng, Prajjwal Bhargava, Pratik Dubal, Praveen Krishnan, Punit~Singh Koura, Puxin Xu, Qing He, Qingxiao Dong, Ragavan Srinivasan, Raj Ganapathy, Ramon Calderer, Ricardo~Silveira Cabral, Robert Stojnic, Roberta Raileanu, Rohit Girdhar, Rohit Patel, Romain Sauvestre, Ronnie Polidoro, Roshan Sumbaly, Ross Taylor, Ruan Silva, Rui Hou, Rui Wang, Saghar Hosseini, Sahana Chennabasappa, Sanjay Singh, Sean Bell, Seohyun~Sonia Kim, Sergey Edunov, Shaoliang Nie, Sharan Narang, Sharath Raparthy, Sheng Shen, Shengye Wan, Shruti Bhosale, Shun Zhang, Simon Vandenhende, Soumya Batra, Spencer Whitman, Sten Sootla, Stephane Collot, Suchin Gururangan, Sydney Borodinsky, Tamar Herman, Tara Fowler, Tarek Sheasha, Thomas Georgiou, Thomas Scialom, Tobias Speckbacher, Todor Mihaylov, Tong Xiao, Ujjwal Karn, Vedanuj Goswami, Vibhor Gupta, Vignesh Ramanathan, Viktor Kerkez, Vincent Gonguet, Virginie Do, Vish Vogeti, Vladan Petrovic, Weiwei Chu,
  Wenhan Xiong, Wenyin Fu, Whitney Meers, Xavier Martinet, Xiaodong Wang, Xiaoqing~Ellen Tan, Xinfeng Xie, Xuchao Jia, Xuewei Wang, Yaelle Goldschlag, Yashesh Gaur, Yasmine Babaei, Yi~Wen, Yiwen Song, Yuchen Zhang, Yue Li, Yuning Mao, Zacharie~Delpierre Coudert, Zheng Yan, Zhengxing Chen, Zoe Papakipos, Aaditya Singh, Aaron Grattafiori, Abha Jain, Adam Kelsey, Adam Shajnfeld, Adithya Gangidi, Adolfo Victoria, Ahuva Goldstand, Ajay Menon, Ajay Sharma, Alex Boesenberg, Alex Vaughan, Alexei Baevski, Allie Feinstein, Amanda Kallet, Amit Sangani, Anam Yunus, Andrei Lupu, Andres Alvarado, Andrew Caples, Andrew Gu, Andrew Ho, Andrew Poulton, Andrew Ryan, Ankit Ramchandani, Annie Franco, Aparajita Saraf, Arkabandhu Chowdhury, Ashley Gabriel, Ashwin Bharambe, Assaf Eisenman, Azadeh Yazdan, Beau James, Ben Maurer, Benjamin Leonhardi, Bernie Huang, Beth Loyd, Beto~De Paola, Bhargavi Paranjape, Bing Liu, Bo~Wu, Boyu Ni, Braden Hancock, Bram Wasti, Brandon Spence, Brani Stojkovic, Brian Gamido, Britt Montalvo, Carl
  Parker, Carly Burton, Catalina Mejia, Changhan Wang, Changkyu Kim, Chao Zhou, Chester Hu, Ching-Hsiang Chu, Chris Cai, Chris Tindal, Christoph Feichtenhofer, Damon Civin, Dana Beaty, Daniel Kreymer, Daniel Li, Danny Wyatt, David Adkins, David Xu, Davide Testuggine, Delia David, Devi Parikh, Diana Liskovich, Didem Foss, Dingkang Wang, Duc Le, Dustin Holland, Edward Dowling, Eissa Jamil, Elaine Montgomery, Eleonora Presani, Emily Hahn, Emily Wood, Erik Brinkman, Esteban Arcaute, Evan Dunbar, Evan Smothers, Fei Sun, Felix Kreuk, Feng Tian, Firat Ozgenel, Francesco Caggioni, Francisco Guzmán, Frank Kanayet, Frank Seide, Gabriela~Medina Florez, Gabriella Schwarz, Gada Badeer, Georgia Swee, Gil Halpern, Govind Thattai, Grant Herman, Grigory Sizov, Guangyi, Zhang, Guna Lakshminarayanan, Hamid Shojanazeri, Han Zou, Hannah Wang, Hanwen Zha, Haroun Habeeb, Harrison Rudolph, Helen Suk, Henry Aspegren, Hunter Goldman, Ibrahim Damlaj, Igor Molybog, Igor Tufanov, Irina-Elena Veliche, Itai Gat, Jake Weissman, James
  Geboski, James Kohli, Japhet Asher, Jean-Baptiste Gaya, Jeff Marcus, Jeff Tang, Jennifer Chan, Jenny Zhen, Jeremy Reizenstein, Jeremy Teboul, Jessica Zhong, Jian Jin, Jingyi Yang, Joe Cummings, Jon Carvill, Jon Shepard, Jonathan McPhie, Jonathan Torres, Josh Ginsburg, Junjie Wang, Kai Wu, Kam~Hou U, Karan Saxena, Karthik Prasad, Kartikay Khandelwal, Katayoun Zand, Kathy Matosich, Kaushik Veeraraghavan, Kelly Michelena, Keqian Li, Kun Huang, Kunal Chawla, Kushal Lakhotia, Kyle Huang, Lailin Chen, Lakshya Garg, Lavender A, Leandro Silva, Lee Bell, Lei Zhang, Liangpeng Guo, Licheng Yu, Liron Moshkovich, Luca Wehrstedt, Madian Khabsa, Manav Avalani, Manish Bhatt, Maria Tsimpoukelli, Martynas Mankus, Matan Hasson, Matthew Lennie, Matthias Reso, Maxim Groshev, Maxim Naumov, Maya Lathi, Meghan Keneally, Michael~L. Seltzer, Michal Valko, Michelle Restrepo, Mihir Patel, Mik Vyatskov, Mikayel Samvelyan, Mike Clark, Mike Macey, Mike Wang, Miquel~Jubert Hermoso, Mo~Metanat, Mohammad Rastegari, Munish Bansal, Nandhini
  Santhanam, Natascha Parks, Natasha White, Navyata Bawa, Nayan Singhal, Nick Egebo, Nicolas Usunier, Nikolay~Pavlovich Laptev, Ning Dong, Ning Zhang, Norman Cheng, Oleg Chernoguz, Olivia Hart, Omkar Salpekar, Ozlem Kalinli, Parkin Kent, Parth Parekh, Paul Saab, Pavan Balaji, Pedro Rittner, Philip Bontrager, Pierre Roux, Piotr Dollar, Polina Zvyagina, Prashant Ratanchandani, Pritish Yuvraj, Qian Liang, Rachad Alao, Rachel Rodriguez, Rafi Ayub, Raghotham Murthy, Raghu Nayani, Rahul Mitra, Raymond Li, Rebekkah Hogan, Robin Battey, Rocky Wang, Rohan Maheswari, Russ Howes, Ruty Rinott, Sai~Jayesh Bondu, Samyak Datta, Sara Chugh, Sara Hunt, Sargun Dhillon, Sasha Sidorov, Satadru Pan, Saurabh Verma, Seiji Yamamoto, Sharadh Ramaswamy, Shaun Lindsay, Shaun Lindsay, Sheng Feng, Shenghao Lin, Shengxin~Cindy Zha, Shiva Shankar, Shuqiang Zhang, Shuqiang Zhang, Sinong Wang, Sneha Agarwal, Soji Sajuyigbe, Soumith Chintala, Stephanie Max, Stephen Chen, Steve Kehoe, Steve Satterfield, Sudarshan Govindaprasad, Sumit Gupta,
  Sungmin Cho, Sunny Virk, Suraj Subramanian, Sy~Choudhury, Sydney Goldman, Tal Remez, Tamar Glaser, Tamara Best, Thilo Kohler, Thomas Robinson, Tianhe Li, Tianjun Zhang, Tim Matthews, Timothy Chou, Tzook Shaked, Varun Vontimitta, Victoria Ajayi, Victoria Montanez, Vijai Mohan, Vinay~Satish Kumar, Vishal Mangla, Vítor Albiero, Vlad Ionescu, Vlad Poenaru, Vlad~Tiberiu Mihailescu, Vladimir Ivanov, Wei Li, Wenchen Wang, Wenwen Jiang, Wes Bouaziz, Will Constable, Xiaocheng Tang, Xiaofang Wang, Xiaojian Wu, Xiaolan Wang, Xide Xia, Xilun Wu, Xinbo Gao, Yanjun Chen, Ye~Hu, Ye~Jia, Ye~Qi, Yenda Li, Yilin Zhang, Ying Zhang, Yossi Adi, Youngjin Nam, Yu, Wang, Yuchen Hao, Yundi Qian, Yuzi He, Zach Rait, Zachary DeVito, Zef Rosnbrick, Zhaoduo Wen, Zhenyu Yang, and Zhiwei Zhao. 2024.
\newblock \href {http://arxiv.org/abs/2407.21783} {The llama 3 herd of models}.

\bibitem[{Eikema and Aziz(2020)}]{eikema-aziz-2020-map}
Bryan Eikema and Wilker Aziz. 2020.
\newblock \href {https://doi.org/10.18653/v1/2020.coling-main.398} {Is {MAP} decoding all you need? the inadequacy of the mode in neural machine translation}.
\newblock In \emph{Proceedings of the 28th International Conference on Computational Linguistics}, pages 4506--4520, Barcelona, Spain (Online). International Committee on Computational Linguistics.

\bibitem[{Fernandes et~al.(2022)Fernandes, Farinhas, Rei, C.~de Souza, Ogayo, Neubig, and Martins}]{fernandes-etal-2022-quality}
Patrick Fernandes, Ant{\'o}nio Farinhas, Ricardo Rei, Jos{\'e}~G. C.~de Souza, Perez Ogayo, Graham Neubig, and Andre Martins. 2022.
\newblock \href {https://doi.org/10.18653/v1/2022.naacl-main.100} {Quality-aware decoding for neural machine translation}.
\newblock In \emph{Proceedings of the 2022 Conference of the North American Chapter of the Association for Computational Linguistics: Human Language Technologies}, pages 1396--1412, Seattle, United States. Association for Computational Linguistics.

\bibitem[{Finkelstein et~al.(2024)Finkelstein, Naskar, Mirzazadeh, Shah, and Freitag}]{finkelstein2024mbrqefinetuningtrainingtime}
Mara Finkelstein, Subhajit Naskar, Mehdi Mirzazadeh, Apurva Shah, and Markus Freitag. 2024.
\newblock \href {http://arxiv.org/abs/2309.10966} {Mbr and qe finetuning: Training-time distillation of the best and most expensive decoding methods}.

\bibitem[{Freitag et~al.(2022{\natexlab{a}})Freitag, Grangier, Tan, and Liang}]{freitag-etal-2022-high}
Markus Freitag, David Grangier, Qijun Tan, and Bowen Liang. 2022{\natexlab{a}}.
\newblock \href {https://doi.org/10.1162/tacl_a_00491} {High quality rather than high model probability: Minimum {B}ayes risk decoding with neural metrics}.
\newblock \emph{Transactions of the Association for Computational Linguistics}, 10:811--825.

\bibitem[{Freitag et~al.(2023)Freitag, Mathur, Lo, Avramidis, Rei, Thompson, Kocmi, Blain, Deutsch, Stewart, Zerva, Castilho, Lavie, and Foster}]{freitag-etal-2023-results}
Markus Freitag, Nitika Mathur, Chi-kiu Lo, Eleftherios Avramidis, Ricardo Rei, Brian Thompson, Tom Kocmi, Frederic Blain, Daniel Deutsch, Craig Stewart, Chrysoula Zerva, Sheila Castilho, Alon Lavie, and George Foster. 2023.
\newblock \href {https://doi.org/10.18653/v1/2023.wmt-1.51} {Results of {WMT}23 metrics shared task: Metrics might be guilty but references are not innocent}.
\newblock In \emph{Proceedings of the Eighth Conference on Machine Translation}, pages 578--628, Singapore. Association for Computational Linguistics.

\bibitem[{Freitag et~al.(2022{\natexlab{b}})Freitag, Rei, Mathur, Lo, Stewart, Avramidis, Kocmi, Foster, Lavie, and Martins}]{freitag-etal-2022-results}
Markus Freitag, Ricardo Rei, Nitika Mathur, Chi-kiu Lo, Craig Stewart, Eleftherios Avramidis, Tom Kocmi, George Foster, Alon Lavie, and Andr{\'e} F.~T. Martins. 2022{\natexlab{b}}.
\newblock \href {https://aclanthology.org/2022.wmt-1.2} {Results of {WMT}22 metrics shared task: Stop using {BLEU} {--} neural metrics are better and more robust}.
\newblock In \emph{Proceedings of the Seventh Conference on Machine Translation (WMT)}, pages 46--68, Abu Dhabi, United Arab Emirates (Hybrid). Association for Computational Linguistics.

\bibitem[{Guerreiro et~al.(2023)Guerreiro, Rei, van Stigt, Coheur, Colombo, and Martins}]{guerreiro2023xcomet}
Nuno~M Guerreiro, Ricardo Rei, Daan van Stigt, Luisa Coheur, Pierre Colombo, and Andr{\'e}~FT Martins. 2023.
\newblock xcomet: Transparent machine translation evaluation through fine-grained error detection.
\newblock \emph{arXiv preprint arXiv:2310.10482}.

\bibitem[{Hendy et~al.(2023)Hendy, Abdelrehim, Sharaf, Raunak, Gabr, Matsushita, Kim, Afify, and Awadalla}]{hendy2023good}
Amr Hendy, Mohamed Abdelrehim, Amr Sharaf, Vikas Raunak, Mohamed Gabr, Hitokazu Matsushita, Young~Jin Kim, Mohamed Afify, and Hany~Hassan Awadalla. 2023.
\newblock How good are gpt models at machine translation? a comprehensive evaluation.
\newblock \emph{arXiv preprint arXiv:2302.09210}.

\bibitem[{Jiao et~al.(2023)Jiao, Wang, Huang, Wang, Shi, and Tu}]{jiao2023chatgpt}
Wenxiang Jiao, Wenxuan Wang, Jen-tse Huang, Xing Wang, Shuming Shi, and Zhaopeng Tu. 2023.
\newblock Is chatgpt a good translator? yes with gpt-4 as the engine.
\newblock \emph{arXiv preprint arXiv:2301.08745}.

\bibitem[{Juraska et~al.(2023)Juraska, Finkelstein, Deutsch, Siddhant, Mirzazadeh, and Freitag}]{juraska-etal-2023-metricx}
Juraj Juraska, Mara Finkelstein, Daniel Deutsch, Aditya Siddhant, Mehdi Mirzazadeh, and Markus Freitag. 2023.
\newblock \href {https://doi.org/10.18653/v1/2023.wmt-1.63} {{M}etric{X}-23: The {G}oogle submission to the {WMT} 2023 metrics shared task}.
\newblock In \emph{Proceedings of the Eighth Conference on Machine Translation}, pages 756--767, Singapore. Association for Computational Linguistics.

\bibitem[{Kocmi et~al.(2023)Kocmi, Avramidis, Bawden, Bojar, Dvorkovich, Federmann, Fishel, Freitag, Gowda, Grundkiewicz, Haddow, Koehn, Marie, Monz, Morishita, Murray, Nagata, Nakazawa, Popel, Popovi{\'c}, and Shmatova}]{kocmi-etal-2023-findings}
Tom Kocmi, Eleftherios Avramidis, Rachel Bawden, Ond{\v{r}}ej Bojar, Anton Dvorkovich, Christian Federmann, Mark Fishel, Markus Freitag, Thamme Gowda, Roman Grundkiewicz, Barry Haddow, Philipp Koehn, Benjamin Marie, Christof Monz, Makoto Morishita, Kenton Murray, Makoto Nagata, Toshiaki Nakazawa, Martin Popel, Maja Popovi{\'c}, and Mariya Shmatova. 2023.
\newblock \href {https://doi.org/10.18653/v1/2023.wmt-1.1} {Findings of the 2023 conference on machine translation ({WMT}23): {LLM}s are here but not quite there yet}.
\newblock In \emph{Proceedings of the Eighth Conference on Machine Translation}, pages 1--42, Singapore. Association for Computational Linguistics.

\bibitem[{Kocmi et~al.(2021)Kocmi, Federmann, Grundkiewicz, Junczys-Dowmunt, Matsushita, and Menezes}]{kocmi-etal-2021-ship}
Tom Kocmi, Christian Federmann, Roman Grundkiewicz, Marcin Junczys-Dowmunt, Hitokazu Matsushita, and Arul Menezes. 2021.
\newblock \href {https://aclanthology.org/2021.wmt-1.57} {To ship or not to ship: An extensive evaluation of automatic metrics for machine translation}.
\newblock In \emph{Proceedings of the Sixth Conference on Machine Translation}, pages 478--494, Online. Association for Computational Linguistics.

\bibitem[{Kocmi et~al.(2024)Kocmi, Zouhar, Federmann, and Post}]{kocmi2024navigating}
Tom Kocmi, Vil{\'e}m Zouhar, Christian Federmann, and Matt Post. 2024.
\newblock Navigating the metrics maze: Reconciling score magnitudes and accuracies.
\newblock \emph{arXiv preprint arXiv:2401.06760}.

\bibitem[{Koehn and Knowles(2017)}]{koehn-knowles-2017-six}
Philipp Koehn and Rebecca Knowles. 2017.
\newblock \href {https://doi.org/10.18653/v1/W17-3204} {Six challenges for neural machine translation}.
\newblock In \emph{Proceedings of the First Workshop on Neural Machine Translation}, pages 28--39, Vancouver. Association for Computational Linguistics.

\bibitem[{Kumar and Byrne(2002)}]{kumar2002minimum}
Shankar Kumar and Bill Byrne. 2002.
\newblock Minimum bayes-risk word alignments of bilingual texts.
\newblock In \emph{Proceedings of the 2002 Conference on Empirical Methods in Natural Language Processing (EMNLP 2002)}, pages 140--147.

\bibitem[{Kumar and Byrne(2004)}]{kumar-byrne-2004-minimum}
Shankar Kumar and William Byrne. 2004.
\newblock \href {https://aclanthology.org/N04-1022} {Minimum {B}ayes-risk decoding for statistical machine translation}.
\newblock In \emph{Proceedings of the Human Language Technology Conference of the North {A}merican Chapter of the Association for Computational Linguistics: {HLT}-{NAACL} 2004}, pages 169--176, Boston, Massachusetts, USA. Association for Computational Linguistics.

\bibitem[{Kwon et~al.(2023)Kwon, Li, Zhuang, Sheng, Zheng, Yu, Gonzalez, Zhang, and Stoica}]{kwon2023efficient}
Woosuk Kwon, Zhuohan Li, Siyuan Zhuang, Ying Sheng, Lianmin Zheng, Cody~Hao Yu, Joseph~E. Gonzalez, Hao Zhang, and Ion Stoica. 2023.
\newblock Efficient memory management for large language model serving with pagedattention.
\newblock In \emph{Proceedings of the ACM SIGOPS 29th Symposium on Operating Systems Principles}.

\bibitem[{Lin(2004)}]{lin-2004-rouge}
Chin-Yew Lin. 2004.
\newblock \href {https://aclanthology.org/W04-1013} {{ROUGE}: A package for automatic evaluation of summaries}.
\newblock In \emph{Text Summarization Branches Out}, pages 74--81, Barcelona, Spain. Association for Computational Linguistics.

\bibitem[{Luong and Manning(2015)}]{luong-manning-2015-stanford}
Minh-Thang Luong and Christopher Manning. 2015.
\newblock \href {https://aclanthology.org/2015.iwslt-evaluation.11} {{S}tanford neural machine translation systems for spoken language domains}.
\newblock In \emph{Proceedings of the 12th International Workshop on Spoken Language Translation: Evaluation Campaign}, pages 76--79, Da Nang, Vietnam.

\bibitem[{Mathur et~al.(2020)Mathur, Baldwin, and Cohn}]{mathur-etal-2020-tangled}
Nitika Mathur, Timothy Baldwin, and Trevor Cohn. 2020.
\newblock \href {https://doi.org/10.18653/v1/2020.acl-main.448} {Tangled up in {BLEU}: Reevaluating the evaluation of automatic machine translation evaluation metrics}.
\newblock In \emph{Proceedings of the 58th Annual Meeting of the Association for Computational Linguistics}, pages 4984--4997, Online. Association for Computational Linguistics.

\bibitem[{OpenAI(2023)}]{openai2023gpt}
R~OpenAI. 2023.
\newblock Gpt-4 technical report. arxiv 2303.08774.
\newblock \emph{View in Article}, 2(5).

\bibitem[{Ott et~al.(2018)Ott, Auli, Grangier, and Ranzato}]{pmlr-v80-ott18a}
Myle Ott, Michael Auli, David Grangier, and Marc'Aurelio Ranzato. 2018.
\newblock \href {https://proceedings.mlr.press/v80/ott18a.html} {Analyzing uncertainty in neural machine translation}.
\newblock In \emph{Proceedings of the 35th International Conference on Machine Learning}, volume~80 of \emph{Proceedings of Machine Learning Research}, pages 3956--3965. PMLR.

\bibitem[{Papineni et~al.(2002)Papineni, Roukos, Ward, and Zhu}]{papineni2002bleu}
Kishore Papineni, Salim Roukos, Todd Ward, and Wei-Jing Zhu. 2002.
\newblock Bleu: a method for automatic evaluation of machine translation.
\newblock In \emph{Proceedings of the 40th annual meeting of the Association for Computational Linguistics}, pages 311--318.

\bibitem[{Popovi{\'c}(2015)}]{popovic2015chrf}
Maja Popovi{\'c}. 2015.
\newblock chrf: character n-gram f-score for automatic mt evaluation.
\newblock In \emph{Proceedings of the tenth workshop on statistical machine translation}, pages 392--395.

\bibitem[{Rafailov et~al.(2024)Rafailov, Sharma, Mitchell, Manning, Ermon, and Finn}]{rafailov2024direct}
Rafael Rafailov, Archit Sharma, Eric Mitchell, Christopher~D Manning, Stefano Ermon, and Chelsea Finn. 2024.
\newblock Direct preference optimization: Your language model is secretly a reward model.
\newblock \emph{Advances in Neural Information Processing Systems}, 36.

\bibitem[{Rei et~al.(2022{\natexlab{a}})Rei, C.~de Souza, Alves, Zerva, Farinha, Glushkova, Lavie, Coheur, and Martins}]{rei-etal-2022-comet}
Ricardo Rei, Jos{\'e}~G. C.~de Souza, Duarte Alves, Chrysoula Zerva, Ana~C Farinha, Taisiya Glushkova, Alon Lavie, Luisa Coheur, and Andr{\'e} F.~T. Martins. 2022{\natexlab{a}}.
\newblock \href {https://aclanthology.org/2022.wmt-1.52} {{COMET}-22: Unbabel-{IST} 2022 submission for the metrics shared task}.
\newblock In \emph{Proceedings of the Seventh Conference on Machine Translation (WMT)}, pages 578--585, Abu Dhabi, United Arab Emirates (Hybrid). Association for Computational Linguistics.

\bibitem[{Rei et~al.(2023)Rei, Guerreiro, Pombal, van Stigt, Treviso, Coheur, C.~de Souza, and Martins}]{rei-etal-2023-scaling}
Ricardo Rei, Nuno~M. Guerreiro, Jos{\~A}{\copyright} Pombal, Daan van Stigt, Marcos Treviso, Luisa Coheur, Jos{\'e}~G. C.~de Souza, and Andr{\'e} Martins. 2023.
\newblock \href {https://doi.org/10.18653/v1/2023.wmt-1.73} {Scaling up {C}omet{K}iwi: Unbabel-{IST} 2023 submission for the quality estimation shared task}.
\newblock In \emph{Proceedings of the Eighth Conference on Machine Translation}, pages 841--848, Singapore. Association for Computational Linguistics.

\bibitem[{Rei et~al.(2020)Rei, Stewart, Farinha, and Lavie}]{rei-etal-2020-comet}
Ricardo Rei, Craig Stewart, Ana~C Farinha, and Alon Lavie. 2020.
\newblock \href {https://doi.org/10.18653/v1/2020.emnlp-main.213} {{COMET}: A neural framework for {MT} evaluation}.
\newblock In \emph{Proceedings of the 2020 Conference on Empirical Methods in Natural Language Processing (EMNLP)}, pages 2685--2702, Online. Association for Computational Linguistics.

\bibitem[{Rei et~al.(2022{\natexlab{b}})Rei, Treviso, Guerreiro, Zerva, Farinha, Maroti, C.~de Souza, Glushkova, Alves, Coheur, Lavie, and Martins}]{rei-etal-2022-cometkiwi}
Ricardo Rei, Marcos Treviso, Nuno~M. Guerreiro, Chrysoula Zerva, Ana~C Farinha, Christine Maroti, Jos{\'e}~G. C.~de Souza, Taisiya Glushkova, Duarte Alves, Luisa Coheur, Alon Lavie, and Andr{\'e} F.~T. Martins. 2022{\natexlab{b}}.
\newblock \href {https://aclanthology.org/2022.wmt-1.60} {{C}omet{K}iwi: {IST}-unbabel 2022 submission for the quality estimation shared task}.
\newblock In \emph{Proceedings of the Seventh Conference on Machine Translation (WMT)}, pages 634--645, Abu Dhabi, United Arab Emirates (Hybrid). Association for Computational Linguistics.

\bibitem[{Sellam et~al.(2020)Sellam, Das, and Parikh}]{sellam-etal-2020-bleurt}
Thibault Sellam, Dipanjan Das, and Ankur Parikh. 2020.
\newblock \href {https://doi.org/10.18653/v1/2020.acl-main.704} {{BLEURT}: Learning robust metrics for text generation}.
\newblock In \emph{Proceedings of the 58th Annual Meeting of the Association for Computational Linguistics}, pages 7881--7892, Online. Association for Computational Linguistics.

\bibitem[{Sennrich et~al.(2016)Sennrich, Haddow, and Birch}]{sennrich-etal-2016-improving}
Rico Sennrich, Barry Haddow, and Alexandra Birch. 2016.
\newblock \href {https://doi.org/10.18653/v1/P16-1009} {Improving neural machine translation models with monolingual data}.
\newblock In \emph{Proceedings of the 54th Annual Meeting of the Association for Computational Linguistics (Volume 1: Long Papers)}, pages 86--96, Berlin, Germany. Association for Computational Linguistics.

\bibitem[{Simianer(2018)}]{simianer2018preference}
Patrick Simianer. 2018.
\newblock \emph{Preference Learning for Machine Translation}.
\newblock Ph.D. thesis.

\bibitem[{Team et~al.(2022)Team, Costa-juss{\`a}, Cross, {\c{C}}elebi, Elbayad, Heafield, Heffernan, Kalbassi, Lam, Licht et~al.}]{nllb2022no}
NLLB Team, Marta~R Costa-juss{\`a}, James Cross, Onur {\c{C}}elebi, Maha Elbayad, Kenneth Heafield, Kevin Heffernan, Elahe Kalbassi, Janice Lam, Daniel Licht, et~al. 2022.
\newblock No language left behind: Scaling human-centered machine translation (2022).
\newblock \emph{URL https://arxiv. org/abs/2207.04672}.

\bibitem[{Wu et~al.(2024)Wu, Nagata, Miao, and Tsuruoka}]{wu2024word}
Qiyu Wu, Masaaki Nagata, Zhongtao Miao, and Yoshimasa Tsuruoka. 2024.
\newblock Word alignment as preference for machine translation.
\newblock \emph{arXiv preprint arXiv:2405.09223}.

\bibitem[{Xu et~al.(2023)Xu, Kim, Sharaf, and Awadalla}]{xu2023paradigm}
Haoran Xu, Young~Jin Kim, Amr Sharaf, and Hany~Hassan Awadalla. 2023.
\newblock A paradigm shift in machine translation: Boosting translation performance of large language models.
\newblock \emph{arXiv preprint arXiv:2309.11674}.

\bibitem[{Xu et~al.(2024{\natexlab{a}})Xu, Sharaf, Chen, Tan, Shen, Van~Durme, Murray, and Kim}]{xu2024contrastive}
Haoran Xu, Amr Sharaf, Yunmo Chen, Weiting Tan, Lingfeng Shen, Benjamin Van~Durme, Kenton Murray, and Young~Jin Kim. 2024{\natexlab{a}}.
\newblock Contrastive preference optimization: Pushing the boundaries of llm performance in machine translation.
\newblock \emph{arXiv preprint arXiv:2401.08417}.

\bibitem[{Xu et~al.(2024{\natexlab{b}})Xu, Zhao, Zu, Gui, Zhang, and Huang}]{xu2024advancing}
Nuo Xu, Jun Zhao, Can Zu, Tao Gui, Qi~Zhang, and Xuanjing Huang. 2024{\natexlab{b}}.
\newblock Advancing translation preference modeling with rlhf: A step towards cost-effective solution.
\newblock \emph{arXiv preprint arXiv:2402.11525}.

\bibitem[{Yan et~al.(2023)Yan, Wang, Zhao, Huang, Chen, and Wang}]{yan-etal-2023-bleurt}
Yiming Yan, Tao Wang, Chengqi Zhao, Shujian Huang, Jiajun Chen, and Mingxuan Wang. 2023.
\newblock \href {https://doi.org/10.18653/v1/2023.acl-long.297} {{BLEURT} has universal translations: An analysis of automatic metrics by minimum risk training}.
\newblock In \emph{Proceedings of the 61st Annual Meeting of the Association for Computational Linguistics (Volume 1: Long Papers)}, pages 5428--5443, Toronto, Canada. Association for Computational Linguistics.

\bibitem[{Yang et~al.(2023)Yang, Chen, Lin, and Byrne}]{yang2023direct}
Guangyu Yang, Jinghong Chen, Weizhe Lin, and Bill Byrne. 2023.
\newblock Direct preference optimization for neural machine translation with minimum bayes risk decoding.
\newblock \emph{arXiv preprint arXiv:2311.08380}.

\bibitem[{Yuan et~al.(2024)Yuan, Pang, Cho, Li, Sukhbaatar, Xu, and Weston}]{yuan2024selfrewardinglanguagemodels}
Weizhe Yuan, Richard~Yuanzhe Pang, Kyunghyun Cho, Xian Li, Sainbayar Sukhbaatar, Jing Xu, and Jason Weston. 2024.
\newblock \href {http://arxiv.org/abs/2401.10020} {Self-rewarding language models}.

\bibitem[{Zhu et~al.(2023)Zhu, Liu, Dong, Xu, Huang, Kong, Chen, and Li}]{zhu2023multilingual}
Wenhao Zhu, Hongyi Liu, Qingxiu Dong, Jingjing Xu, Shujian Huang, Lingpeng Kong, Jiajun Chen, and Lei Li. 2023.
\newblock Multilingual machine translation with large language models: Empirical results and analysis.
\newblock \emph{arXiv preprint arXiv:2304.04675}.

\end{thebibliography}
\bibliographystyle{acl_natbib}

\appendix

\clearpage

\onecolumn

\section{Additional Results}
\label{apdx:results}

In this section, we present results on WMT'23 test data. The findings in Tables \ref{tab:results_wmt23}, \ref{tab:no_sys_xps_wmt23}, \ref{tab:chosen_sys_impact_wmt23} and \ref{tab:mono_vs_multi_wmt23} support the observations discussed in the main text for the WMT'22 dataset. In Tables \ref{tab:wmt22_lps} and \ref{tab:wmt23_lps}, we also provide additional insights, split by language pairs, and include extra metrics, specifically Metric-X and BLEU.

\begin{table*}[h]
\centering
\small
\resizebox{0.88\textwidth}{!}
{
\begin{tabular}{lccccccccc}
\toprule
 & \multicolumn{4}{c}{\texttt{xx-en}} &  & \multicolumn{4}{c}{\texttt{en-xx}} \\
 \noalign{\vskip 0.5pt}
\cline{2-5}
\cline{7-10}
\noalign{\vskip 2pt}
 & \multicolumn{2}{c}{Neural} &  & Lexical &  & \multicolumn{2}{c}{Neural} &  & Lexical \\
 \noalign{\vskip 0.5pt}
\cline{2-3}
\cline{5-5}
\cline{7-8}
\cline{10-10}
\noalign{\vskip 2pt}
 & xCOMET-QE & CometKiwi &  & chrF &  & xCOMET-QE & CometKiwi &  & chrF \\
\midrule
\textbf{Base} & \textcolor{white}{$\bullet$} \colorcell{88.00}{88.00} & \textcolor{white}{$\bullet$} \colorcell{77.74}{77.74} &  & \textcolor{white}{$\bullet$} \colorcell{52.30}{52.30} &  & \textcolor{white}{$\bullet$} \colorcell{86.19}{86.19} & \textcolor{white}{$\bullet$} \colorcell{73.08}{73.08} &  & \textcolor{white}{$\bullet$} \colorcell{47.31}{47.31} \\
\midrule
\multicolumn{10}{c}{\textit{Preferences induced with} \textbf{xCOMET-QE}} \\\cdashlinelr{1-10}
SFT & \textcolor{atomictangerine!60}{$\bullet$} \colorcell{88.96}{88.00} & \textcolor{white}{$\bullet$} \colorcell{78.46}{77.74} &  & \textcolor{white}{$\bullet$} \colorcell{53.30}{52.30} &  & \textcolor{atomictangerine!60}{$\bullet$} \colorcell{87.07}{86.19} & \textcolor{white}{$\bullet$} \colorcell{73.99}{73.08} &  & \textcolor{white}{$\bullet$} \colorcell{48.38}{47.31} \\
CPO & \textcolor{atomictangerine!60}{$\bullet$} \textit{\colorcell{89.77}{88.00}} & \textcolor{white}{$\bullet$} \textit{\colorcell{78.95}{77.74}} &  & \textcolor{white}{$\bullet$} \colorcell{53.47}{52.30} &  & \textcolor{atomictangerine!60}{$\bullet$} \textit{\colorcell{88.09}{86.19}} & \textcolor{white}{$\bullet$} \textit{\colorcell{76.75}{73.08}} &  & \textcolor{white}{$\bullet$} \textit{\colorcell{44.29}{47.31}} \\
\midrule
\multicolumn{10}{c}{\textit{Preferences induced with} \textbf{CometKiwi}} \\
\bottomrule
\end{tabular}}
\caption{Comparison between SFT on preferred translations and CPO in the multi-system setting on WMT'23 test data. Notations and formatting are the same as in Table~\ref{tab:results}.}
\label{tab:results_wmt23}
\end{table*}

\begin{table*}[h]
\centering
\small
\resizebox{0.88\textwidth}{!}
{\begin{tabular}{lccccccccc}
\toprule
  & \multicolumn{4}{c}{\texttt{xx-en}} &  & \multicolumn{4}{c}{\texttt{en-xx}} \\
\noalign{\vskip 0.5pt}
\cline{2-5}
\cline{7-10}
\noalign{\vskip 2pt}
 & \multicolumn{2}{c}{Neural} &  & Lexical &  & \multicolumn{2}{c}{Neural} &  & Lexical \\
 \noalign{\vskip 0.5pt}
\cline{2-3}
\cline{5-5}
\cline{7-8}
\cline{10-10}
\noalign{\vskip 2pt}
 & xCOMET-QE & CometKiwi &  & chrF &  & xCOMET-QE & CometKiwi &  & chrF \\
\midrule
\textbf{Base} & \textcolor{white}{$\bullet$} \colorcell{88.00}{88.00} & \textcolor{white}{$\bullet$} \colorcell{77.74}{77.74} &  & \textcolor{white}{$\bullet$} \colorcell{52.30}{52.30} &  & \textcolor{white}{$\bullet$} \colorcell{86.19}{86.19} & \textcolor{white}{$\bullet$} \colorcell{73.08}{73.08} &  & \textcolor{white}{$\bullet$} \colorcell{47.31}{47.31} \\\midrule
\multicolumn{10}{c}{\textit{Optimization via} \textbf{SFT}} \\\cdashlinelr{1-10}
\multicolumn{10}{l}{\textit{Preferences induced with} \textbf{xCOMET-QE}}\\
All systems & \textcolor{atomictangerine!60}{$\bullet$} \colorcell{88.96}{88.00} & \textcolor{white}{$\bullet$} \colorcell{78.46}{77.74} &  & \textcolor{white}{$\bullet$} \colorcell{53.30}{52.30} &  & \textcolor{atomictangerine!60}{$\bullet$} \colorcell{87.07}{86.19} & \textcolor{white}{$\bullet$} \colorcell{73.99}{73.08} &  & \textcolor{white}{$\bullet$} \colorcell{48.38}{47.31} \\
No Base & \textcolor{atomictangerine!60}{$\bullet$} \colorcell{89.07}{88.00} & \textcolor{white}{$\bullet$} \colorcell{78.53}{77.74} &  & \textcolor{white}{$\bullet$} \textit{\colorcell{53.57}{52.30}} &  & \textcolor{atomictangerine!60}{$\bullet$} \textit{\colorcell{86.94}{86.19}} & \textcolor{white}{$\bullet$} \textit{\colorcell{73.70}{73.08}} &  & \textcolor{white}{$\bullet$} \textit{\colorcell{48.52}{47.31}} \\
\bottomrule
\end{tabular}}
\caption{Impact of candidate systems on WMT'23 downstream performance in the multi-system setting. Notations and formatting are the same as in Table~\ref{tab:no_sys_xps}.}
\label{tab:no_sys_xps_wmt23}
\end{table*}

\clearpage

\begin{table*}[h]
\centering
\small
\resizebox{0.88\textwidth}{!}
{\begin{tabular}{lccccccccc}
\toprule
 & \multicolumn{4}{c}{\texttt{xx-en}} &  & \multicolumn{4}{c}{\texttt{en-xx}} \\
\noalign{\vskip 0.5pt}
\cline{2-5}
\cline{7-10}
\noalign{\vskip 2pt}
 & \multicolumn{2}{c}{Neural} &  & Lexical &  & \multicolumn{2}{c}{Neural} &  & Lexical \\
 \noalign{\vskip 0.5pt}
\cline{2-3}
\cline{5-5}
\cline{7-8}
\cline{10-10}
\noalign{\vskip 2pt}
 & xCOMET-QE & CometKiwi &  & chrF &  & xCOMET-QE & CometKiwi &  & chrF \\
\midrule
\textbf{Base} & \textcolor{white}{$\bullet$} \colorcell{88.00}{88.00} & \textcolor{white}{$\bullet$} \colorcell{77.74}{77.74} &  & \textcolor{white}{$\bullet$} \colorcell{52.30}{52.30} &  & \textcolor{white}{$\bullet$} \colorcell{86.19}{86.19} & \textcolor{white}{$\bullet$} \colorcell{73.08}{73.08} &  & \textcolor{white}{$\bullet$} \colorcell{47.31}{47.31} \\\midrule
\multicolumn{10}{c}{\textit{Chosen system set to} \textbf{Base}} \\\cdashlinelr{1-10}
SFT & \textcolor{atomictangerine!60}{$\bullet$} \colorcell{88.07}{88.00} & \textcolor{white}{$\bullet$} \colorcell{77.93}{77.74} &  & \textcolor{white}{$\bullet$} \colorcell{52.52}{52.30} &  & \textcolor{atomictangerine!60}{$\bullet$} \colorcell{86.52}{86.19} & \textcolor{white}{$\bullet$} \colorcell{73.27}{73.08} &  & \textcolor{white}{$\bullet$} \colorcell{47.52}{47.31} \\
CPO & \textcolor{atomictangerine!60}{$\bullet$} \colorcell{88.05}{88.00} & \textcolor{white}{$\bullet$} \colorcell{77.95}{77.74} &  & \textcolor{white}{$\bullet$} \textit{\colorcell{52.24}{52.30}} &  & \textcolor{atomictangerine!60}{$\bullet$} \textit{\colorcell{86.68}{86.19}} & \textcolor{white}{$\bullet$} \textit{\colorcell{73.75}{73.08}} &  & \textcolor{white}{$\bullet$} \textit{\colorcell{46.54}{47.31}} \\\midrule
\multicolumn{10}{c}{\textit{Chosen system set to} \textbf{Ref}} \\
\bottomrule
\end{tabular}}
\caption{Impact of the chosen system on WMT'23 downstream performance in the multi-system setting. Notations and formatting are the same as in Table~\ref{tab:chosen_sys_impact}.}
\label{tab:chosen_sys_impact_wmt23}
\end{table*}

\begin{table*}[h]
\centering
\small
\resizebox{0.88\textwidth}{!}
{\begin{tabular}{lccccccccc}
\toprule
 & \multicolumn{4}{c}{\texttt{xx-en}} &  & \multicolumn{4}{c}{\texttt{en-xx}} \\
 \noalign{\vskip 0.5pt}
\cline{2-5}
\cline{7-10}
\noalign{\vskip 2pt}
 & \multicolumn{2}{c}{Neural} &  & Lexical &  & \multicolumn{2}{c}{Neural} &  & Lexical \\
 \noalign{\vskip 0.5pt}
\cline{2-3}
\cline{5-5}
\cline{7-8}
\cline{10-10}
\noalign{\vskip 2pt}
 & xCOMET-QE & CometKiwi &  & chrF &  & xCOMET-QE & CometKiwi &  & chrF \\
\midrule
\textbf{Base} & \textcolor{white}{$\bullet$} \colorcell{88.00}{88.00} & \textcolor{white}{$\bullet$} \colorcell{77.74}{77.74} &  & \textcolor{white}{$\bullet$} \colorcell{52.30}{52.30} &  & \textcolor{white}{$\bullet$} \colorcell{86.19}{86.19} & \textcolor{white}{$\bullet$} \colorcell{73.08}{73.08} &  & \textcolor{white}{$\bullet$} \colorcell{47.31}{47.31} \\\midrule
\multicolumn{10}{c}{\textit{Optimization via} \textbf{SFT}} \\\cdashlinelr{1-10}
Multi-system & \textcolor{atomictangerine!60}{$\bullet$} \colorcell{88.96}{88.00} & \textcolor{white}{$\bullet$} \colorcell{78.46}{77.74} &  & \textcolor{white}{$\bullet$} \colorcell{53.30}{52.30} &  & \textcolor{atomictangerine!60}{$\bullet$} \colorcell{87.07}{86.19} & \textcolor{white}{$\bullet$} \colorcell{73.99}{73.08} &  & \textcolor{white}{$\bullet$} \colorcell{48.38}{47.31} \\
Mono-system & \textcolor{atomictangerine!60}{$\bullet$} \textit{\colorcell{88.55}{88.00}} & \textcolor{white}{$\bullet$} \textit{\colorcell{78.17}{77.74}} &  & \textcolor{white}{$\bullet$} \textit{\colorcell{52.74}{52.30}} &  & \textcolor{atomictangerine!60}{$\bullet$}\textit{ \colorcell{86.75}{86.19}} & \textcolor{white}{$\bullet$} \colorcell{73.87}{73.08} &  & \textcolor{white}{$\bullet$} \textit{\colorcell{47.43}{47.31}} \\\midrule
\multicolumn{10}{c}{\textit{Optimization via} \textbf{CPO}} \\
\bottomrule
\end{tabular}}
\caption{Comparison between multi- and mono-system fine-tuning on WMT'23 test data. Notations and formatting are the same as in Table~\ref{tab:mono_vs_multi}.}
\label{tab:mono_vs_multi_wmt23}
\end{table*}

\begin{table*}[h]
\centering
\small
\resizebox{0.99\textwidth}{!}
{\begin{tabular}{lccccccccccccccccccccccccccc}
\toprule
 & \multicolumn{6}{c}{cs-en} &  & \multicolumn{6}{c}{en-cs} &  & \multicolumn{6}{c}{de-en} &  & \multicolumn{6}{c}{en-de} \\
 \noalign{\vskip 0.5pt}
\cline{2-7}
\cline{9-14}
\cline{16-21}
\cline{23-28}
\noalign{\vskip 2pt}
 & \multicolumn{3}{c}{Neural} &  & \multicolumn{2}{c}{Lexical} &  & \multicolumn{3}{c}{Neural} &  & \multicolumn{2}{c}{Lexical} &  & \multicolumn{3}{c}{Neural} &  & \multicolumn{2}{c}{Lexical} &  & \multicolumn{3}{c}{Neural} &  & \multicolumn{2}{c}{Lexical} \\
 \noalign{\vskip 0.5pt}
\cline{2-4}
\cline{6-7}
\cline{9-11}
\cline{13-14}
\cline{16-18}
\cline{20-21}
\cline{23-25}
\cline{27-28}
\noalign{\vskip 2pt}
 & xCOMET-QE & CometKiwi & Metric-X &  & chrF & BLEU &  & xCOMET-QE & CometKiwi & Metric-X &  & chrF & BLEU &  & xCOMET-QE & CometKiwi & Metric-X &  & chrF & BLEU &  & xCOMET-QE & CometKiwi & Metric-X &  & chrF & BLEU \\
\midrule

\textbf{Base} & 83.42 & 82.57 & 2.00 &  & 65.81 & 41.25 &  & 90.80 & 81.96 & 1.48 &  & 53.58 & 27.42 &  & 93.33 & 83.48 & 2.03 &  & 55.24 & 29.02 &  & 96.32 & 80.87 & 1.22 &  & 56.95 & 27.65 \\

\noalign{\vskip 0.5pt}\cline{1-28}\noalign{\vskip 2pt}
\textbf{SFT} &&&&&&&&&&&&&&&&&&&&&&&&&&& \\
\textit{Multi-system} &&&&&&&&&&&&&&&&&&&&&&&&&&& \\
$\rightarrow$ xCOMET-QE &&&&&&&&&&&&&&&&&&&&&&&&&&& \\

Vanilla & 86.18 & 83.17 & 1.98 &  & 67.36 & 42.81 &  & 91.49 & 82.69 & 1.42 &  & 55.02 & 28.58 &  & 93.88 & 83.85 & 2.00 &  & 56.32 & 29.86 &  & 96.57 & 81.21 & 1.19 &  & 57.62 & 28.22 \\
\bottomrule
\end{tabular}}
\caption{Comprehensive downstream evaluation for the WMT'22 dataset, reporting xCOMET-QE, CometKiwi, Metric-X, chrF, and BLEU scores for all models and language pairs. Learning objectives are indicated in \textbf{bold} font, candidate settings in \textit{italics}, and alignment metrics are preceded by an arrow ($\rightarrow$).}
\label{tab:wmt22_lps}
\end{table*}

\begin{table*}[h]
\centering
\small
\resizebox{0.9\textwidth}{!}
{\begin{tabular}{lcccccccccccccccccccc}
\toprule
 & \multicolumn{6}{c}{en-cs} &  & \multicolumn{6}{c}{de-en} &  & \multicolumn{6}{c}{en-de} \\
 \noalign{\vskip 0.5pt}
\cline{2-7}
\cline{9-14}
\cline{16-21}
\noalign{\vskip 2pt}
 & \multicolumn{3}{c}{Neural} &  & \multicolumn{2}{c}{Lexical} &  & \multicolumn{3}{c}{Neural} &  & \multicolumn{2}{c}{Lexical} &  & \multicolumn{3}{c}{Neural} &  & \multicolumn{2}{c}{Lexical} \\
 \noalign{\vskip 0.5pt}
\cline{2-4}
\cline{6-7}
\cline{9-11}
\cline{13-14}
\cline{16-18}
\cline{20-21}
\noalign{\vskip 2pt}
 & xCOMET-QE & CometKiwi & Metric-X &  & chrF & BLEU &  & xCOMET-QE & CometKiwi & Metric-X &  & chrF & BLEU &  & xCOMET-QE & CometKiwi & Metric-X &  & chrF & BLEU \\
\midrule
\textbf{Base} & 85.90 & 73.23 & 1.91 &  & 52.57 & 27.45 &  & 84.79 & 76.57 & 3.73 &  & 66.64 & 39.38 &  & 84.97 & 71.97 & 3.04 &  & 61.69 & 32.78 \\

\noalign{\vskip 0.5pt}\cline{1-21}\noalign{\vskip 2pt}
\textbf{SFT} &&&&&&&&&&&&&&&&&&&& \\
\textit{Multi-system} &&&&&&&&&&&&&&&&&&&& \\
$\rightarrow$ xCOMET-QE &&&&&&&&&&&&&&&&&&&& \\

Vanilla & 87.19 & 74.32 & 1.78 &  & 54.31 & 28.71 &  & 85.29 & 77.01 & 3.59 &  & 67.82 & 40.50 &  & 85.75 & 72.56 & 2.91 &  & 61.75 & 32.81 \\
\bottomrule
\end{tabular}}
\caption{Comprehensive downstream evaluation for the WMT'23 dataset. Metrics, notations and formatting are the same as in Table~\ref{tab:wmt22_lps}.}
\label{tab:wmt23_lps}
\end{table*}

\clearpage

\section{Additional Data Details}
\label{apdx:data_details}

\subsection{Building Preference Datasets in the Mono-System Setting}
\label{apdx:pref_build}

Following the experimental setup detailed in the main text (Section~\ref{sec:xp_setup}), we here provide further details on the method used to construct mono-system preference datasets. As a reminder, after generating the $K$ candidate translations for each source sentence, we have, for all $1 \leq i \leq N$,
\begin{equation*}
\begin{split}
\mathcal{Y}_i^{mono} = \left\{ y_i^{1}, \cdots, y_i^{K} \right\}
\end{split},
\end{equation*}
where $y_i^{1} \preceq \cdots \preceq y_i^{K}$ are assumed to be sorted in increasing metric score order. For each sample, we evaluate $y_i^{Base}$ (the greedy-decoded translation) using metric $m$ and check its rank in the set of candidate translations. We denote it by $b_i$. Sorted in increasing quality order, we thereby have 
\begin{equation*}
\begin{split}
y_i^{1} \preceq \cdots \preceq y_i^{b_i-1} \preceq y_i^{Base} \preceq y_i^{b_i} \preceq \cdots \preceq y_i^{K}
\end{split}.
\end{equation*}
Finally, to determine the chosen and rejected hypotheses, we select two offset parameters $o^r, o^c \in \mathbb{N}$, such that the chosen and rejected options are respectively 
\begin{equation*}
\begin{split}
    \begin{cases}
        \begin{aligned}
            y_i^c = y_i^{\min(K, b_i + o^c)} \\
            y_i^r = y_i^{\max(1, b_i - o^r)} 
        \end{aligned}
    \end{cases}.
\end{split}
\end{equation*}
Intuitively, $o^r$ and $o^c$ control the average quality of the chosen and rejected options in the resulting preference dataset and ensure that the chosen (resp. rejected) option always has a higher (resp. lower) quality than the base translation. Table~\ref{tab:data_stats} presents the average quality properties for mono-system preference datasets, and compares them to the multi-system setting.

\begin{table*}[h]
\centering
\small
\resizebox{0.65\textwidth}{!}
{\begin{tabular}{llcccc}
\toprule
 &  & \multicolumn{2}{c}{Neural} &  & Lexical \\
\noalign{\vskip 0.5pt}
\cline{3-4}
\cline{6-6}
\noalign{\vskip 2pt}
 & Hyp. & xCOMET-QE & CometKiwi &  & chrF \\
\midrule
\textbf{Multi-system} &&&&& \\
\multirow{3}{*}{Candidate systems} & Base & 93.09 & 87.13 &  & 58.33 \\
 & GPT-4 & 94.58 & 88.32 &  & 60.93 \\
 & Reference & 91.84 & 86.72 &  & 100.00 \\
\noalign{\vskip 0.5pt}\cdashline{1-6}[0.5pt/2pt]\noalign{\vskip 2pt}
\multirow{2}{*}{Vanilla preference dataset} & Rejected & 87.86 & 84.15 &  & 78.48 \\
 & Chosen & 97.24 & 89.81 &  & 75.95 \\
\noalign{\vskip 0.5pt}\cdashline{1-6}\noalign{\vskip 2pt}
\textbf{Mono-system} &&&&& \\
\multirow{2}{*}{Multi-system replica} & Rejected & 87.80 & 83.04 &  & 55.69 \\
 & Chosen & 97.29 & 89.20 &  & 57.18 \\
\noalign{\vskip 0.5pt}\cdashline{1-6}[0.5pt/2pt]\noalign{\vskip 2pt}
\multirow{2}{*}{Chosen = Low / Rejected = Low} & Rejected & 75.36 & 75.46 &  & 52.95 \\
 & Chosen & 93.60 & 87.04 &  & 57.14 \\
\noalign{\vskip 0.5pt}\cdashline{1-6}[0.5pt/2pt]\noalign{\vskip 2pt}
\multirow{2}{*}{Chosen = Low / Rejected = Mid} & Rejected & 84.54 & 81.02 &  & 54.93 \\
 & Chosen & 93.60 & 87.04 &  & 57.14 \\
\noalign{\vskip 0.5pt}\cdashline{1-6}[0.5pt/2pt]\noalign{\vskip 2pt}
\multirow{2}{*}{Chosen = Low / Rejected = High} & Rejected & 92.15 & 85.54 &  & 55.86 \\
 & Chosen & 93.60 & 87.04 &  & 57.14 \\
\noalign{\vskip 0.5pt}\cdashline{1-6}[0.5pt/2pt]\noalign{\vskip 2pt}
\multirow{2}{*}{Chosen = Mid / Rejected = Low} & Rejected & 75.36 & 75.46 &  & 52.95 \\
 & Chosen & 95.77 & 88.40 &  & 57.43 \\
\noalign{\vskip 0.5pt}\cdashline{1-6}[0.5pt/2pt]\noalign{\vskip 2pt}
\multirow{2}{*}{Chosen = Mid / Rejected = Mid} & Rejected & 84.54 & 81.02 &  & 54.93 \\
 & Chosen & 95.77 & 88.40 &  & 57.43 \\
\noalign{\vskip 0.5pt}\cdashline{1-6}[0.5pt/2pt]\noalign{\vskip 2pt}
\multirow{2}{*}{Chosen = Mid / Rejected = High} & Rejected & 92.15 & 85.54 &  & 55.86 \\
 & Chosen & 95.77 & 88.40 &  & 57.43 \\
\noalign{\vskip 0.5pt}\cdashline{1-6}[0.5pt/2pt]\noalign{\vskip 2pt}
\multirow{2}{*}{Chosen = High / Rejected = Low} & Rejected & 75.36 & 75.46 &  & 52.95 \\
 & Chosen & 98.16 & 89.84 &  & 57.56 \\
\noalign{\vskip 0.5pt}\cdashline{1-6}[0.5pt/2pt]\noalign{\vskip 2pt}
\multirow{2}{*}{Chosen = High / Rejected = Mid} & Rejected & 84.54 & 81.02 &  & 54.93 \\
 & Chosen & 98.16 & 89.84 &  & 57.56 \\
\noalign{\vskip 0.5pt}\cdashline{1-6}[0.5pt/2pt]\noalign{\vskip 2pt}
\multirow{2}{*}{Chosen = High / Rejected = High} & Rejected & 92.15 & 85.54 &  & 55.86 \\
 & Chosen & 98.16 & 89.84 &  & 57.56 \\
\bottomrule
\end{tabular}}
\caption{Average quality properties for xCOMET-QE-based mono-system preference datasets, compared to the multi-system setting. Multi-system replica is the mono-system dataset that matches the average chosen/rejected qualities of the multi-system preference data. Other mono-system datasets are represented by their relative average chosen/rejected qualities.}
\label{tab:data_stats}
\end{table*}

\clearpage

\subsection{Language Statistics}

\begin{figure*}[h]
  \centering
  \includegraphics[width=0.75\textwidth]{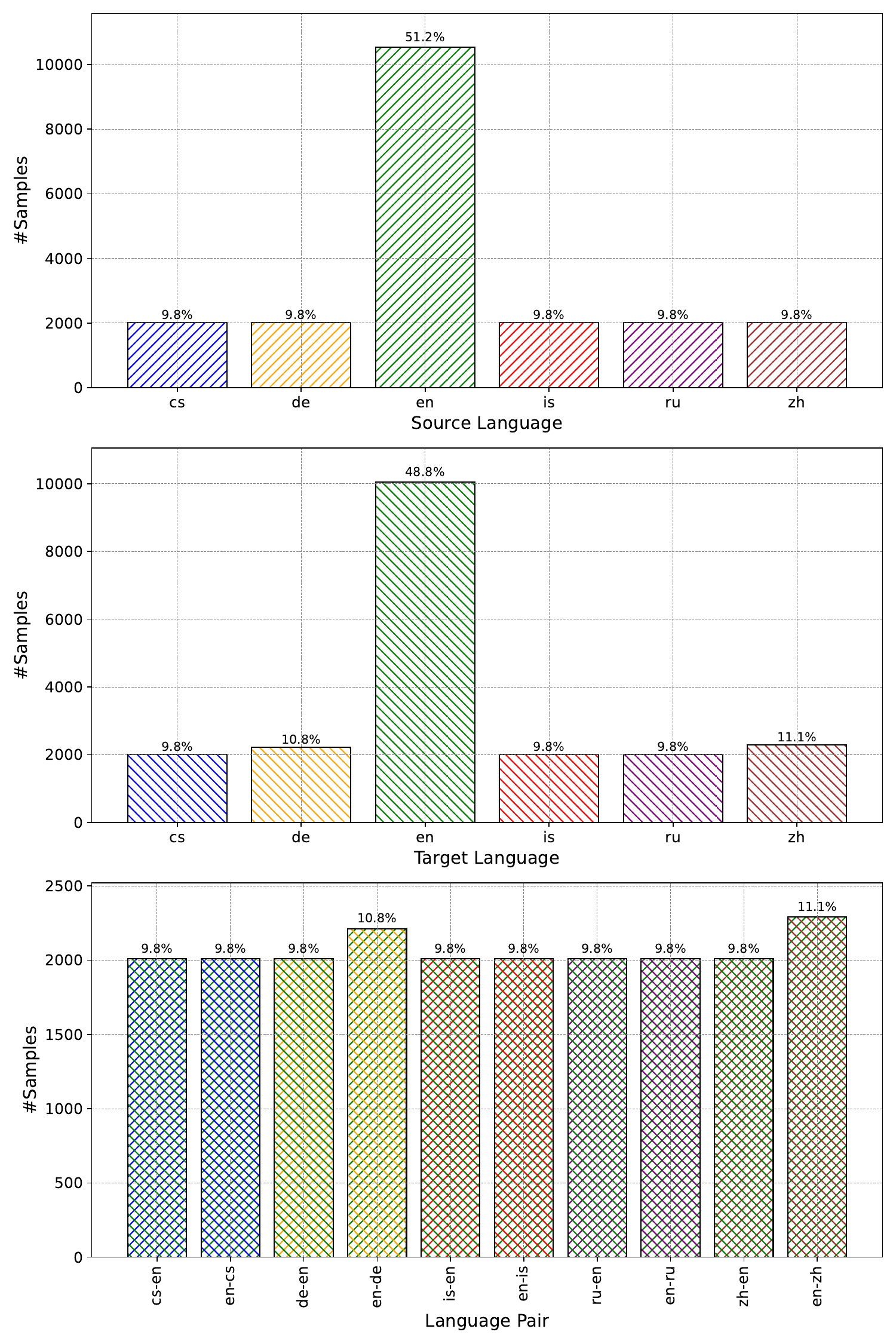}
  \caption{Language statistics for preference datasets. The y-axis represents the number of samples, corresponding percentages are displayed above each bar.}
  \label{apdx:ft_langs}
\end{figure*}

\begin{figure*}[t]
  \centering
  \includegraphics[width=0.9\textwidth]{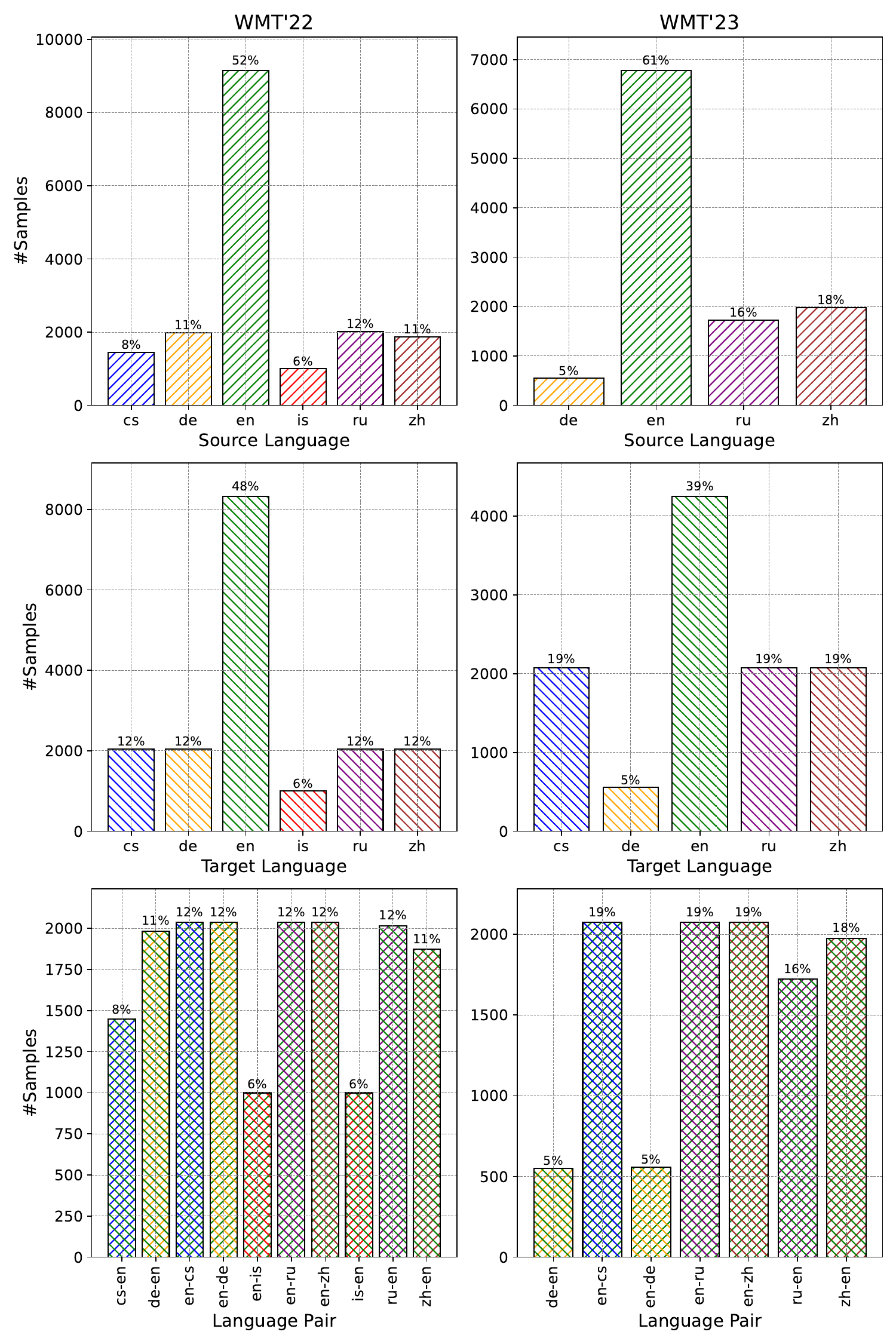}
  \caption{Language statistics for WMT'22 and WMT'23 test data. The y-axis represents the number of samples, corresponding percentages are displayed above each bar.}
  \label{apdx:test_langs}
\end{figure*}

\end{document}